\documentclass[entropy,article,accept,moreauthors,pdftex]{Definitions/mdpi} 
\graphicspath{{Figures/}}
\firstpage{1} 
\pubvolume{1}
\issuenum{1}
\articlenumber{0}
\pubyear{2021}
\copyrightyear{2020}
\datereceived{} 
\dateaccepted{} 
\datepublished{} 
\hreflink{https://doi.org/} 

\usepackage{graphicx}
\usepackage{subcaption}
\usepackage{amsmath}

\DeclareMathOperator{\EX}{\mathbb{E}}

\Title{Conditional Deep Gaussian Processes: Empirical Bayes Hyperdata Learning}
\TitleCitation{Conditional DGP: empirical Bayes hyperdata learning}

\Author{Chi-Ken Lu $^{1,*}$\orcidA{} and Patrick Shafto $^{1,2}$}

\AuthorNames{Chi-Ken Lu and Patrick Shafto}

\AuthorCitation{Lu, C.-K.; Shafto, P.}

\address{%
$^{1}$ \quad Mathematics and Computer Science, Rutgers University, Newark, NJ 07102 \\
$^{2}$ \quad School of Mathematics, Institute for Advanced Studies, Princeton 
}

\corres{Correspondence: cl1178@rutgers.edu
}

\abstract{
It is desirable to combine the expressive power of deep learning with Gaussian Process (GP) in one expressive Bayesian learning model. Deep kernel learning 
showed success in adopting a deep network for feature extraction followed by a GP used as function model. Recently, 
it was suggested that, albeit training with marginal likelihood, the deterministic nature of feature extractor might lead to overfitting while the replacement with a Bayesian network seemed to cure it. Here, we propose the conditional Deep Gaussian Process (DGP) in which the intermediate GPs in hierarchical composition are supported by the hyperdata and the exposed GP remains zero mean. Motivated by the inducing points in sparse GP, the hyperdata also play the role of function supports, but are hyperparameters rather than random variables. We follow our previous moment matching approach
to approximate the marginal prior for conditional DGP with a GP carrying an effective kernel. Thus, as in empirical Bayes, the hyperdata are learned by optimizing the approximate marginal likelihood which implicitly depends on the hyperdata via the kernel. We shall show the equivalence with the deep kernel learning in the limit of dense hyperdata in latent space. However, the conditional DGP and the corresponding approximate inference enjoy the benefit of being more Bayesian than deep kernel learning. Preliminary extrapolation results demonstrate expressive power from the depth of hierarchy by exploiting the exact covariance and hyperdata learning, in comparison with  
GP kernel composition, DGP variational inference and deep kernel learning. 
We also address the non-Gaussian aspect of our model as well as way of upgrading to a full Bayes inference.   
}


\begin{document}

\section{Introduction}

Deep Gaussian Process~\cite{damianou2013deep} is a Bayesian learning model which combines both the expressive power of deep neural networks~\cite{goodfellow2016deep} and calibrated uncertainty estimation.   
The hierarchical composition of Gaussian Processes (GPs)~\cite{rasmussen2006gaussian} is the origin of expressiveness, but also renders inference intractable as the marginalization of GPs in the stage of computing evidence is not analytically possible. Expectation propagation~\cite{minka2001expectation,bui2016deep} and variational inference~\cite{salimbeni2017doubly,salimbeni2019deep,haibin2019implicit,ustyuzhaninov2020compositional} are approximate inference schemes for DGP. The latter has issues of posterior collapse, which turns DGP into a GP with transformed input. Ref.~\cite{haibin2019implicit,ustyuzhaninov2020compositional} address this issue and compositional freedom~\cite{havasi2018inference} in such hierarchical learning. 
Nevertheless, inferential challenges continue to slow adoption of DGP. 

Despite challengers, there has been progresses in understanding this seemingly simple yet profound model. In the case where the GPs in the hierarchy are zero-mean, DGP exhibits pathology, becoming a constant function as the depth increases \cite{duvenaud2014avoiding}. Using the fact that the exponential covariance function is strictly convex, \cite{dunlop2018deep} and \cite{tong2021characterizing} studied the conditional statistics for squared distance in function space, suggesting region in hyperparameter space to avoid the pathology. 
Recently, \cite{agrawal2020wide} showed the connection between DGP and a deep neural network with bottlenecked layers, and \cite{pleiss2021limitations} suggested that the DGP with large width may collapse back to GP.

Others have found ways to work around the challenges of DGPs.
The deep kernel learning proposed in~\cite{wilson2016deep} gained the Bayesian character of GP and the expressive power of deep neural network without encountering intractability as the learning of weight parameters, treated as kernel hyperparameters, is an empirical Bayes. 
Similar ideas also appeared in~\cite{salakhutdinov2007using} and \cite{calandra2016manifold}. Hyperparameter learning in~\cite{wilson2016deep} is through the marginal likelihood, which can in principle prevent overfitting due to the built-in competition between data fitting and model complexity~\cite{rasmussen2006gaussian}. However, \cite{ober2021promises} suggested that the lack of Bayesian character in the deep feature extracting net might still result in overfitting if the network has too many parameters.   

Here, we propose a conditional DGP model in which the intermediate GPs (all but the exposed GP) in the hierarchical composition are conditioned on a set of hyperdata. These hyperdata are inspired by the inducing points in sparse GP~\cite{titsias2009variational,titsias2010bayesian,matthews2016sparse}, but {\it they are hyperparameters, not random variables}. 
The conditional DGP is motivated by the expressive power and Bayesian character of DGP~\cite{damianou2013deep} as well as the deep kernel learning with objective in marginal likelihood~\cite{wilson2016deep}. Due to the conditioning on the hyperdata, the intermediate GPs can be viewed as collection of random feature functions centered around the deterministic conditional mean. Thus, the intermediate GPs become approximately deterministic functions when the hyperdata are sufficiently dense. Besides, lifting the intermeidate GPs from being zero mean might help avoid pathology too.
Mathematically, we define a marginal prior for the conditional DGP, i.e. all intermediate GPs are marginalized, which assures the Bayesian character in dealing with the feature functions. We then use the moment matching method to approximate the non-Gaussian marginal prior as a GP~\cite{lu2020interpretable}, which connects with observed data and allows the marginal likelihood objective. It should be stressed that the effective kernel depends on the conditional mean and conditional covariance in feature function via the hyperdata, which are optimized in the spirit of empirical Bayes~\cite{murphy2012machine}.    
In the implementation, the hyperdata supporting each intermediate GP are represented as a neural network function, $u={\rm nn}_{\bf w}(z)$ with $u$ and $z$ being the output and input of hyperdata, similar with the trick used in modeling the mean and variance for data in the variational autoencoder~\cite{kingma2013auto}. 

The paper is organized as follows. Sec. 2 gives a short survey of current literature in deep probabilistic models, usage of moment matching in approximate inference, and the inducing points in GP and DGP. A background of mathematical models of GP and DGP, the marginal prior for DGP, and the moment matching method are introduced in Sec. 3. The conditional DGP with SE kernel in the exposed layer, its mathematical connection with deep kernel learning, the parameter learning, and the non-Gaussian aspect, are described in Sec. 4. Preliminary demonstration on extrapolating two time-series data is in Sec. 5, followed by a discussion in Sec. 6.

\section{Related work}

In the literature on deep probabilistic models, \cite{garnelo2018conditional} proposed Conditional Neural Process
in which the mean and variance functions are learned from the encoded representation of context data in a regression setup for target data.
Deep Gaussian Processes (DGPs) constitute one family of models for composition functions by conditioning input to GP on output of another GP~\cite{damianou2013deep}. Similar idea appeared in the works of warped GP
\cite{snelson2004warped,lazaro2012bayesian}. The Implicit Process in~\cite{ma2019variational} is a stochastic process embedding the Gaussian distribution into a neural network. Solutions of stochastic differential equation driven by GP are also examples of composite process~\cite{ustyuzhaninov2020monotonic}.
Variational DGP
casts inference problem in terms of optimizing ELBO~\cite{salimbeni2017doubly} or EP~\cite{bui2016deep}. However, the multi-modalness of DGP posterior~\cite{havasi2018inference,lu2020interpretable} may arise from the fact that the hidden mappings in intermediate layers are dependent~\cite{ustyuzhaninov2020compositional}. Inference schemes capable of capturing the multi-modal nature of DGP posterior was recently proposed by~\cite{haibin2019implicit,ustyuzhaninov2020compositional}. Depth of neural network models and the function expressivity were studied in~\cite{telgarsky2016benefits,pearce2020expressive}, and uncertainty estimates were investigated in~\cite{gal2016dropout}. DGP in weight space representation and its variational Bayesian approach to DGP inference was introduced in~\cite{cutajar2017random}, which was based on the notion of random feature expansion of Gaussian~\cite{rahimi2008random} and arcsine~\cite{cho2009kernel} kernels. Deep hierarchical SVMs and PCAs were introduced in~\cite{scholkopf1998nonlinear}.

Moment matching is a way to approximate a complex distribution with, for instance, a Gaussian by capturing the mean and the second moment. \cite{girard2003gaussian} considered a GP regression with uncertain input, and replaced the non-Gaussian predictive distribution with a Gaussian carrying the matched mean and variance. Expectation Propagation in \cite{minka2001expectation} computed the vector of mean and variance parameters of non-Gaussian posterior distributions.
\cite{titsias2010bayesian} approximated the distribution over unseen pixels as multivariate Gaussian with matched mean and covariance. Moment matching is also extensively applicable in comparing two distributions~\cite{muandet2012learning} where the embedded means in RKHS are computed. In generative models, the model parameters are learned from comparing the model and data distributions~\cite{li2015generative}.

Inducing points are an important technique in sparse GP~\cite{quinonero2005unifying,titsias2009variational,matthews2016sparse,shi2019sparse} and DGP. In addition to being locally defined as function's input and output, \cite{lazaro2009inter} introduced a transformation to form a global set of inducing {\it features}. One popular transformation uses the basis of Gaussian so that one can recover the local inducing points easily~\cite{lazaro2009inter}. Transformation using the basis of spherical harmonic functions in~\cite{dutordoir2020sparse} allows orthogonal inducing features and connects with the arcsine kernels of Bayesian deep neural network~\cite{dutordoir2021deep}. \cite{rudner2020inter} employed the inter-domain features in DGP inference. Recently, \cite{ober2021global} proposed a method to express the local inducing points in the weight space representation. All the methods cited here treated the inducing points or features in a full Bayes approach as they are random variables, associated with an approximate distribution~\cite{murphy2012machine}.    
\section{Background}
Here, we briefly introduce the notions of Gaussian Process as a model for random continuous function $f({\bf x}):{\mathbb  R^d \mapsto\mathbb R}$. Deep Gaussian Process~\cite{damianou2013deep} is a hierarchical composition of Gaussian Process for modeling general composite function ${\bf f}_L\circ{\bf f}_{L-1}\circ\cdots{\bf f}_2\circ{\bf f}_1({\bf x})$ where the bold faced function ${\bf f}_1:{\mathbb R^d}\mapsto{\mathbb R^{H_1}}$ has a output consisting of $H_1$ independent GPs, and similarly for ${\bf f}_2:\mathbb R^{H_1}\mapsto\mathbb R^{H_2}$ and so on. The depth and width of DGP are thus denoted by $L$ and $H_{1:L}$, respectively.
\subsection{Gaussian Process}
In machine learning, the attention is often restricted to the finite set of correlated random variables ${\bf f}:=\{f({\bf x}_1),\cdots,f({\bf x}_N)\}$ 
corresponding to the design location ${\bf X}=({\bf x}_1,\cdots,{\bf x}_N)^T$. 
Denoting $f_i:=f({\bf x}_i)$, the above set of random variables is a GP if and only if the following relations,
\begin{equation}
    \EX[f_i]=\mu({\bf x}_i)\:,\ \EX[(f_i-\mu_i)(f_j-\mu_j)]=k({\bf x}_i,{\bf x}_j)\:,
\end{equation} are satisfied for all indices $i,j$. For convenience, we can use $f\sim\mathcal{GP}(\mu, k)$ to denote the above. The mean function $\mu(\cdot):\mathbb R^d\mapsto\mathbb R$ and the covariance function $k(\cdot,\cdot):\mathbb R^d\times\mathbb R^d\mapsto\mathbb R$ then fully specify the GP. 
One can proceed to write down the multivariate normal distribution as the pdf, 
\begin{equation}
    p({\bf f}) = \frac{1}{\sqrt{(2\pi)^N|K|}}\exp[-\frac{1}{2}
    ({\bf f-m})^tK^{-1}({\bf f-m})]\:.
\end{equation}
The covariance matrix $K$ has matrix element $K_{ij} = k({\bf x}_i, {\bf x}_j)$, characterizing the correlation between the function values. The covariance function $k$ encodes function properties such as smoothness. The vector ${\bf m}:=\mu({\bf X})$ represents the mean values at corresponding inputs. Popular covariance functions include the squared exponential (SE) $k({\bf x}_i, {\bf x}_j)=\sigma^2\exp[-||{\bf x}_i-{\bf x}_j||^2/(2\ell^2)]$ and the family of Matern functions. The signal magnitude $\sigma$ and length scale $\ell$ are hyper-parameters. 


The conditional property of Gaussians allows one to place constraint on the model $p({\bf f})$. Given a set of function values ${\bf u}=f({\bf Z})$, the space of random function $f$ now only include those passing through these fixed points. Then the conditional pdf $p({\bf f}|{\bf u})$ has the conditional mean and covariance, 
\begin{align}
    {\bf m}&\rightarrow{\bf m} + K_{\bf xz}K^{-1}_{\bf zz}[{\bf u-m}]\label{conditional_mean}\\
    {\bf K_X}&\rightarrow{\bf K_X}-{\bf K}_{\bf XZ}{\bf K}^{-1}_{\bf Z}{\bf K}_{{\bf ZX}}\label{conditional_cov}
\end{align} where the matrix ${\bf K}_{\bf XZ}$ represents the covariance matrix evaluated at ${\bf X}$ against ${\bf Z}$.   

\subsection{Deep Gaussian Process}
We follow the seminal work in~\cite{damianou2013deep} to generalize the notion of GP to the composite functions ${\bf f}_L\circ{\bf f}_{L-1}\circ\cdots{\bf f}_2\circ{\bf f}_1({\bf x})$. 
In most literature, DGP is defined from a generative point of view. Namely, the joint distribution for the simplest zero-mean DGP with $L=2$ and $H_2=H_1=1$ can be expressed as,
\begin{equation}
	p(f_2,f_1|{\bf X})= p(f_2|f_1)p(f_1|{\bf X})\:,
\end{equation} with the conditional defined as $f_2|f_1\sim\mathcal{GP}(0,k(f_1,f_1))$ and $f_1\sim\mathcal{GP}(0,k({\bf X},{\bf X})$. 

\subsection{Marginal prior, covariance, and marginal likelihood}
In the above DGP model, the {\it exposed} GP for $f_2$ is connected with the data output ${\bf y}$ while the {\it intermediate} GP for $f_1$ with the data input ${\bf X}$. In Bayesian learning, both $f$'s shall be marginalized in computing the evidence. Now 
we define the marginal prior as,
\begin{equation}
	p({\bf f})=\int d{\bf f}_1\ p({\bf f}_2|{\bf f}_1)p({\bf f}_1|{\bf X})\:.\label{true_DGP}
\end{equation} in which the bold faced ${\bf f}_1$ representing the set of the intermediate function values are marginalized but the exposed ${\bf f}_2$ is not. Note that the notation $f({\bf x})=f_2(f_1({\bf x}))$ is not ambiguous in a generative view, but may cause some confusion in the marginal view as the label $f_1$ has been integrated out. To avoid confusing with the exposed function $f_2(\cdot)$, we still use $f(\cdot)$ to denote the marginalized composite function unless otherwise stated. 

Motivated to write down an objective in terms of marginal likelihood, the moment matching method in~\cite{lu2020interpretable} was proposed so one can approximate Eq.~(\ref{true_DGP}) with a multivariate Gaussian $q({\bf f|X})$ that the mean and the covariance are matched. In the zero-mean DGP considered in~\cite{lu2020interpretable}, the covariance matching refers to,  
\begin{equation}
	\EX_{{\bf f}\sim q}[f_if_j]=\EX_{{\bf f}_1}[\EX_{{\bf f}_2|{\bf f}_1}[f_if_j]]=
	\int d{\bf f}_2d{\bf f}_1f_2(f_1({\bf x}_i))f_2(f_1({\bf x}_j))p({\bf f}_2|{\bf f}_1)p({\bf f}_1|{\bf X})\:.
\end{equation} In the case where the squared exponential kernel is used in both GPs, the approximate marginal prior $q({\bf f|X})=\mathcal N(0,K_{\rm eff})$ with the effective kernel being $k_{\rm eff}=\sigma_2^2[1+2\frac{\sigma_1^2}{\ell_2^2}(1-\exp(-|{\bf x}_i-{\bf x}_j|^2/2\ell_1^2))]^{-\frac{1}{2}}$~\cite{lu2020interpretable}. The hyperparameters include the length scale $\ell$ and signal magnitude $\sigma$ with layer indexed at the subscript.

Consequently, the evidence of the data ${\bf X, y}$ associated with the 2-layer DGP can be approximately expressed as,
\begin{equation}
	p({\bf y| X})\approx\int d{\bf f} p({\bf y|f})q({\bf f|X})\:.
\end{equation} Thus, the learning of hyperparameters $\sigma$'s and $\ell$'s in the zero-mean DGP model are through the gradient descent on $\log p({\bf y|X})$, and the gradient components $\frac{\partial K}{\partial \ell_{1,2}}$ and $\frac{\partial K}{\partial \sigma_{1,2}}$ are needed in the framework of GPy~\cite{gpy2014}.

\section{Model}
Following the previous discussion, we shall introduce the model of conditional DGP along with the covariance and marginal prior. The mathematical connection with deep kernel learning and the non-Gaussian aspect of marginal prior will be discussed. The difference between the original DGP and the conditional DGP is that the intermediate GPs in latter are conditioned on the hyperdata. Learning the hyperdata via the approximate marginal likelihood is, loosely speaking, an empirical Bayesian learning of the feature function in the setting of deep kernel learning. 

\subsection{Conditional DGP}
In the simple two-layer hierarchy with width $H_1=H_2=1$, the hyperdata $\{{\bf Z, u}\}=\{{\bf z}_{1:M}\in\mathbb R^d, u_{1:M}\in\mathbb R\}$ are introduced as support for the intermediate GP for $f_1$, while the exposed GP for $f_2$ remains zero-mean and does not condition on any point. Thus, $f_1$ can be viewed as a space of random functions constrained with $f_1({\bf z}_{1:M})=u_{1:M}$, and the Gaussian distribution $p(f_1({\bf x}_{1:N})|{\bf Z,u})$ has its conditional mean and covariance in Eq.~(\ref{conditional_mean}) [with ${\bf m}$ on RHS set to zero] and (\ref{conditional_cov}), respectively. Following Eq.~(\ref{true_DGP}), the marginal prior for this conditional DGP can be similarly expressed as,
\begin{equation}
	p({\bf f})=\int d{\bf f}_1 p({\bf f}_2|{\bf f}_1)p({\bf f}_1|{\bf X, Z, u})\:.\label{marginal_prior_cond}
\end{equation} 
With $f_1$ being conditioned on the hyperdata $\{{\bf Z, u}\}$, one can see that the multivariate Gaussian $p(f_1({\bf x}_{1:N})|{\bf Z,u})$ emits samples in the space of random functions passing through the fixed hyperdata so that Eq.~(\ref{marginal_prior_cond}) is a sum of an infinite number of GPs. Namely, 
\begin{equation*}
	f\sim\sum_{f_1}\mathcal{GP}(0,k_2(f_1({\bf X}),f_1(\bf{X})))\:,
\end{equation*} with $f_1$ under the constraints due to the hyperdata and the smoothness implied in kernel $k_1$. Therefore, $f$ are represented by an ensemble of GPs with same kernel but different {\it feature} functions. We shall come back to this point more rigorously in Sec.~\ref{dkl}.

Now we shall approximate the intractable distribution in Eq.~(\ref{marginal_prior_cond}) with a multivariate Gaussian $q({\bf f|X, Z, u})$ carrying the matched covariance. The following lemma is useful for the case where the exposed GP for $f_2|f_1$ uses the squared exponential (SE) kernel.
\begin{Lemma} (Lemma 3 in~\cite{lu2020deep}) The covariance in $p({\bf f})$ [Eq.~(\ref{marginal_prior_cond})] with the SE kernel $k_2(x,y)=\sigma_2^2\exp[-(x-y)^2/2\ell_2^2]$ in the exposed GP for $f_2|f_1$ can be calculated analytically. With the Gaussian conditional distribution, $p({\bf f}_1|{\bf X, Z, u})$, supported by the hyperdata, the effective kernel reads, 
\begin{equation}
    k_{\rm eff}({\bf x}_i,{\bf x}_j)
    =\frac{\sigma_2^2}
    {\sqrt{1+\delta_{ij}^2/\ell_2^2}}\exp\left[-\frac{(m_i-m_j)^2}{2(\ell_2^2+\delta_{ij}^2)}
    \right]\:,\label{kernel:se}
\end{equation} where $m_{i,j}:=m({\bf x}_{i,j})$ and $c_{ij}:={\rm cov}(f_1({\bf x}_i),f_1({\bf x}_j))$ are the conditional mean and covariance, respectively, at the inputs ${\bf x}_{i,j}$. The positive parameter $\delta_{ij}^2:=c_{ii}+c_{jj}-2c_{ij}$ and the the length scale $\ell_2$ dictates how the uncertainty about $f_1$ affects the function composition. 
\end{Lemma}

Next, in addition to the hyperparameters like $\sigma$'s and $\ell$'s, the function values $u_{1:M}$ are hyperdata that shall be learned from the objective. With approximating the non-Gaussian marginal prior $p({\bf f|X, Z, u})$ with $q({\bf f|X,Z,u})$, we are able to compute the approximate marginal likelihood as the objective,
\begin{equation}
	\mathcal L = -\log\int d{\bf f}p({\bf y|f})q({\bf f|X,Z,u})\:.
\end{equation} The learning of all hyperparameters data follow the standard gradient descent used in GP~\cite{gpy2014}, and the gradient components include the usual ones like $\partial{K_{\rm eff}}/\partial {\ell_{2}}$ in exposed layer and those related to the intermediate layer $\partial{K_{\rm eff}}/\partial{\ell_1}$ and the hyperdata $\partial{K_{\rm eff}}/\partial u_{1:M}$ through chaining with $\partial{K_{\rm eff}}/\partial(m_i-m_j)$ and $\partial{K_{\rm eff}}/\partial{\delta^2_{ij}}$ via Eq.~(\ref{conditional_mean}) and (\ref{conditional_cov}). To exploit the expressive power of neural network during optimization, the hyperdata can be further modeled by a neural network, i.e. 
\begin{equation}
	u_{1:M}={\rm nn}_{{\bf w}}({\bf z}_{1:M})\:,
\end{equation} with ${\bf w}$ denoting the weight parameters. In such case, the weights ${\bf w}$ are learned instead of the hyperdata $u_{1:M}$.   
 
\subsection{When conditional DGP is almost a GP}\label{dkl}

In the limiting case where the probabilistic nature of $f_1$ is negligible then the conditional DGP becomes a GP with transformed input, i.e. the distribution $p({\bf f}_1|{\bf X,Z,u})$ becomes highly concentrated around a certain conditional mean ${\bar f_1}({\bf x})$. To get insight, we reexamine the covariance in the setting where $f_1$ is almost deterministic. We can reparameterize the random function $f_1$ at two distinct inputs ${\bf x}_{1,2}$ for the purpose of computing covariance,
\begin{equation}
    f_1({\bf x}_{i,j})=m({\bf x}_{i,j})+\epsilon_{i,j}\:,\label{g_rep}
\end{equation} where $m({\bf x})$ is the conditional mean given the fixed ${\bf Z}$ and ${\bf u}$. The random character lies in the two correlated random variables, $(\epsilon_i,\epsilon_j)^T\sim\mathcal N(0,C)$ corresponding to the weak but correlated signal around zero. Under the assumption, we follow the analysis in \citep{girard2003gaussian,ustyuzhaninov2020compositional} and prove the following lemma. 
\begin{Lemma}
Consider $p(f)$ defined in Eq.~(\ref{marginal_prior_cond}) with $f_2|f_1$ being a more general $\mathcal{GP}(\mu_2,k_2)$ and $f_1$ reparametrized as in Eq.~(\ref{g_rep}). The covariance, ${\rm cov}(f_2(f_1({\bf x}_i)),f_2(f_1({\bf x}_j)))$, has the following form,
\begin{equation}
    [1+\frac{c_{ii}}{2}\partial^2_{m_j}+
    \frac{c_{jj}}{2}\partial^2_{m_i}+
    c_{ij}\partial^2_{m_im_j}]k_2(m_i,m_j)+c_{ij}\mu_2'(m_i)\mu_2'(m_j)\:,\label{total_cov}
\end{equation}
where the $c$'s are matrix elements of kernel matrix $C$ associated with the weak random variables $\epsilon_{i,j}$ in Eq.~(\ref{g_rep}). The notations $m_{i,j}:=m({\bf x}_{i,j})$ and prime as derivative are used.
\end{Lemma}

\begin{proof}
The assumption is that $f_2|f_1\sim\mathcal{GP}(\mu_2,k_2)$ and that $f_1(x)$ a weak random function $\epsilon(x)$ overlaying a fixed function $m(x)$. At any two inputs $x_{i,j}$, we expand the target function $f$ to the second order,
\begin{equation}
    f(x_{i,j})=f_2(f_1(x_{i,j}))\approx f_2(m_{i,j})+\epsilon_{i,j}f_2'(m_{i,j})
    +\frac{\epsilon_{i,j}^2}{2}f_2''(m_{i,j})\:,
\end{equation} where the shorthanded notations $m_i:=m(x_i)$ and $\epsilon(x_i):=\epsilon_i$ are used. Note that $(\epsilon_i,\epsilon_j)$ is bivariate Gaussian with zero mean and covariance matrix ${\bf C}$. We shall use the law of total covariance, ${\rm cov}[a,b]={\rm cov}(\EX[a|d],\EX[b|d])+\EX[{\rm cov}(a,b|d)]$ with $a$, $b$, and $d$ being some random variables. To proceed the first term, we calculate the conditional mean given the $\epsilon$'s,
\begin{equation}
    \EX[f(x_{i,j})|\epsilon_i,\epsilon_j]=\mu_2(m_{i,j})+\epsilon_{i,j}\mu'(m_{i,j})\:.
\end{equation} Then one uses the fact that $f_2|f_1$, $f_2'|f_1$, and $f_2''|f_1$ are jointly Gaussian to compute the conditional covariance, which can be expressed in a compact form,
\begin{equation}
    {\rm cov}[f(x_i)f(x_j)|\epsilon_i,\epsilon_j] = \hat O(\epsilon_i,\epsilon_j)k(m_i,m_j)\:.
\end{equation}
The operator $\hat O$ accounts for the fact that 
${\rm cov}[f_2'(m_i),f_2'(m_j)]=\partial^2_{m_im_j}k_2(m_i,m_j)$ and ${\rm cov}[f_2(m_i),f_2''(m_j)]=\partial^2_{m_j}k_2(m_i,m_j)$. Thus, the operator reads,
\begin{equation}
    \hat O=1+\epsilon_i\partial_{m_i}+\epsilon_j\partial_{m_j} + \frac{\epsilon_i^2}{2}\partial^2_{m_i}+
    \frac{\epsilon_j^2}{2}\partial^2_{m_j}
    +\epsilon_i\epsilon_j\partial^2_{m_im_j}\:.
\end{equation} Now we are ready to deal with the outer expectation with respect to the $\epsilon$'s. Note that the covariance $c_{ij}:=\EX[\epsilon_i\epsilon_j]=c(x_i,x_j)$ and variance $c_{ii}:=\EX[\epsilon_i^2]=c(x_i,x_i)$ are matrix elements of ${\bf C}$. Consequently, we prove the total covariance in Eq.~(\ref{total_cov})
\end{proof}

\begin{Remark} 
Since the second derivatives $\partial^2_{m_i}k_2(m_i,m_j)=\partial^2_{m_j}k_2(m_i,m_j)=-\partial^2_{m_im_j}k_2(m_i,m_j)$ hold for the stationary $k_2$, the above covariance [Eq.~(\ref{total_cov})] with $\mu_2=0$ is identical to the effective kernel in Eq.~(\ref{kernel:se}) in the limit $\ell_2^2\gg\delta_{ij}^2$, which reads
\begin{equation}
    {\rm cov}(f(x_i)f(x_j))\propto
    [1+\frac{(m_i-m_j)^2-\ell_2^2}{2\ell_2^4}\delta_{ij}^2]\exp[-\frac{(m_i-m_j)^2}{2\ell_2^2}]\:.\label{kernel:se_app}
\end{equation}
Such a situation occurs when the inputs ${\bf Z}$ in hyperdata are dense enough so that $f_1$ becomes almost deterministic. 
\end{Remark}



Consequently, in the limit when the conditional covariance in $\delta^2$ term is small compared with the length scale $\ell_2^2$, Eq.~(\ref{kernel:se_app}) indicates that the effective kernel is the SE kernel with a deterministic input $m({\bf x})$, which is equivalent to the deep kernel with SE as the base kernel  [see Eq. (5) in~\cite{wilson2016deep}]. On the other hand, when $\delta^2$ and $\ell_2$ are comparable, the terms within the first bracket in the RHS of Eq.~(\ref{kernel:se_app}) is a non-stationary function which may attribute multiple frequencies in the function $f$. The deep kernel with the spectral mixture kernel [Eq.~(6) in~\cite{wilson2016deep}] as the base is similar with the effective kernel.

\subsection{Non-Gaussian aspect}

The statistics of the non-Gaussian marginal prior $p({\bf f|X,Z,u})$ are not solely determined by the moments up to the second order. The fourth moment can be derived in a similar manner in~\cite{lu2020interpretable} with the help of
the theorem in~\citep{isserlis1918formula}. Relevant discussion of the heavy-tailed character in Bayesian deep neural network can be found in~\cite{vladimirova2019understanding,yaida2020non,zavatone2021exact}. See Lemma A1 for the details of computing the general fourth moment in the case where SE kernel is used in $f_2|f_1$ in the conditional 2-layer DGP. 
Here, we briefly discuss the non-Gaussian aspect, focusing on the variance of covariance, i.e. by comparing $\EX_p[(f({\bf x}_i)f({\bf x}_j))^2]$ and $\EX_q[(f({\bf x}_i)f({\bf x}_j))^2]$ with $p$ being the true distribution [Eq.~(\ref{marginal_prior_cond})] and $q$ being the approximating Gaussian. 


In the SE case, one can verify the difference in the fourth order expectation value,
\begin{equation*}
	\EX_p[(f({\bf x}_i)f({\bf x}_j))^2]-\EX_q[(f({\bf x}_i)f({\bf x}_j))^2]=\frac{e^{-\frac{(m_i-m_j)^2}{1+2\delta_{ij}^2}}}{[1+2\delta_{ij}^2]^{1/2}}
	-\frac{e^{-\frac{(m_i-m_j)^2}{1+\delta_{ij}^2}}}{[1+\delta_{ij}^2]}\geq 0\:,
\end{equation*} where we have used the fact that the inequalities $(1+2\delta^2)^{-1/2}\geq(1+\delta^2)^{-1}$ and $\exp[-(1+2\delta^2)^{-1}]\geq\exp[-(1+\delta^2)^{-1}]$ hold. Therefore, the inequality suggests the heavy-tailed statistics of the marginal prior $p(f({\bf x}_i),f({\bf x}_j))$ over any pair of function values.  





\section{Results}

The works in~\cite{duvenaud2013structure,sun2018differentiable} demonstrate that GPs can still have superior expressive power and generalization if the kernels are dedicatedly designed. With the belief that deeper models generalize better than the shallower counterparts~\cite{mhaskar2016learning}, DGP models are expected to perform better in fitting and generalization than GP models do if the same kernel is used in both. However, such expectation may not be fully realized as the approximate inference may lose some power in DGP. For instance, the diminishing variance in the posterior over the latent function was reported in~\cite{ustyuzhaninov2020compositional} regarding the variational inference for DGP~\cite{salimbeni2017doubly}.  
Here, with demonstration on extrapolating the real-world time series data with the conditional DGP, we shall show that the depth along with optimizing the hyperdata do enhance the expressive power and the generalization due to the multiple length scale and multiple frequencies character of the effective kernel. In addition, the moment matching method as an approximate inference for conditional DGP does not suffer from the posterior collapse. The simulation codes can be found in the \href{https://github.com/luck1226/ConditionalDeepGaussianProcesses}{github repository}.

\subsection{Mauna Loa Data}

\begin{figure}[H]
\widefigure
    \centering
    \begin{subfigure}{4.2 cm}
        \includegraphics[width=\textwidth]{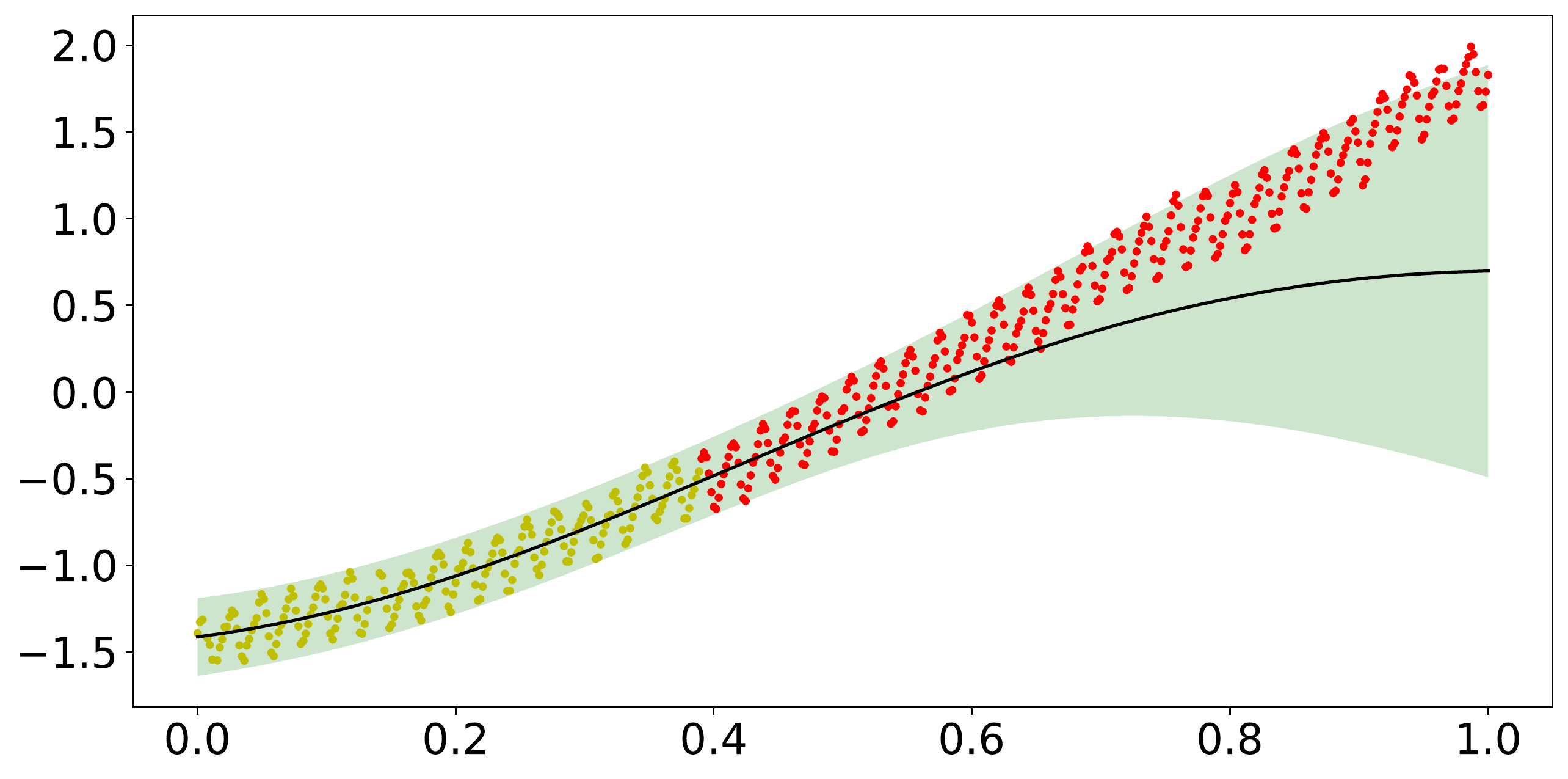}
        \caption{SE kernel}
        \label{se_kernel}
    \end{subfigure}
    ~ 
    \begin{subfigure}{4.2 cm}
        \includegraphics[width=\textwidth]{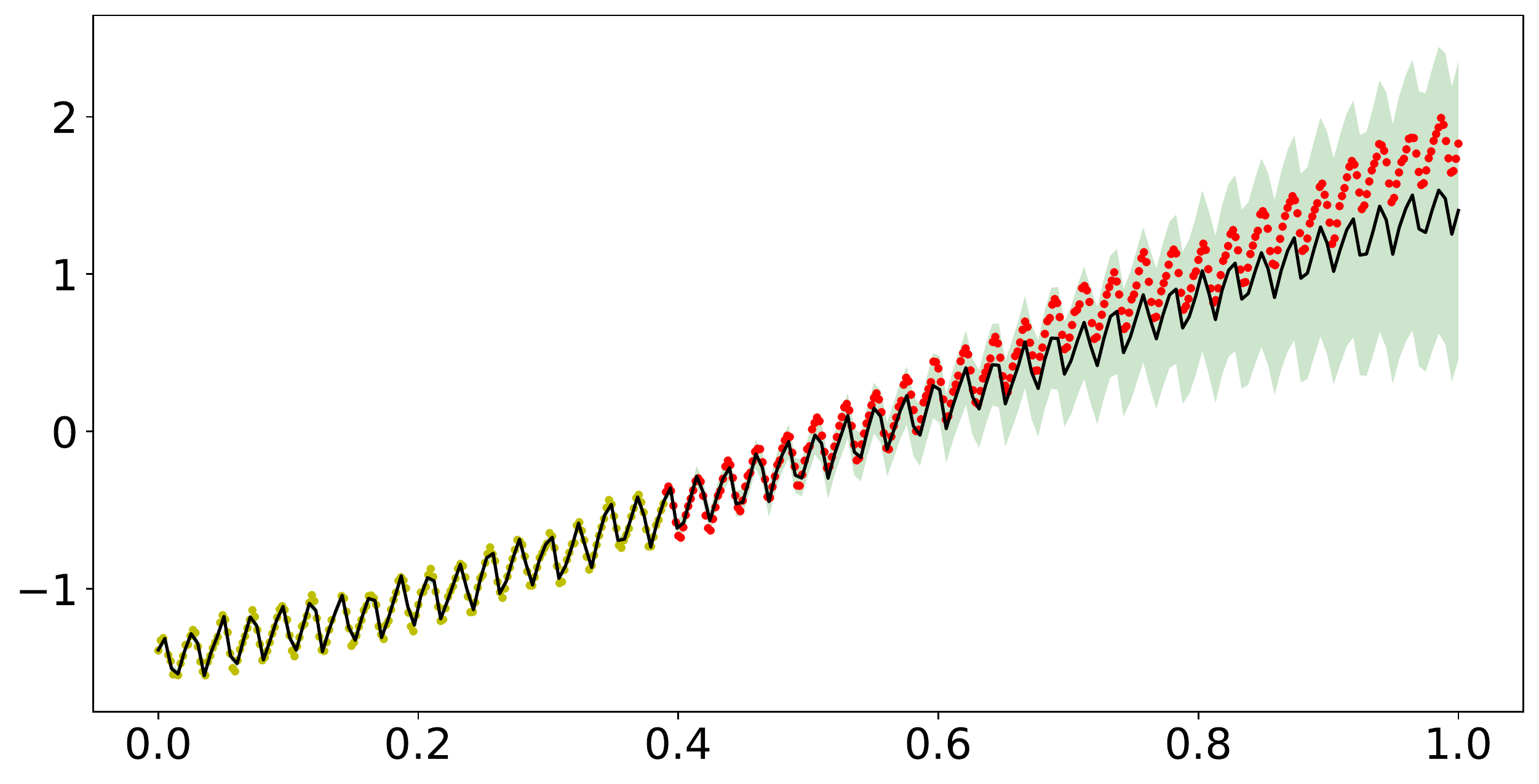}
        \caption{SE+periodic SE+RQ kernel}
        \label{mix_kernel}
    \end{subfigure}
    ~ 
    \begin{subfigure}{4.2 cm}
        \includegraphics[width=\textwidth]{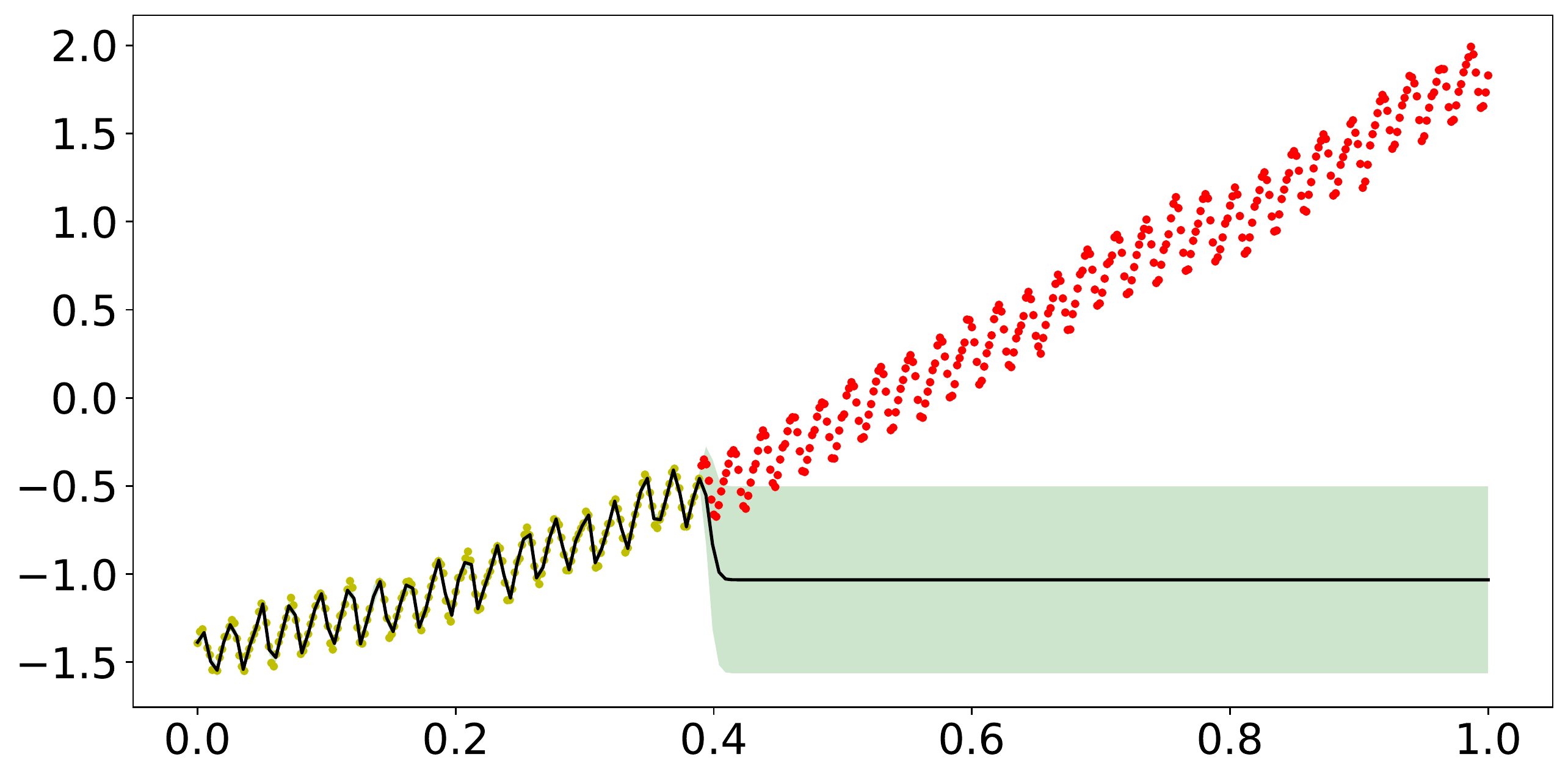}
        \caption{SE[SE] kernel}
        \label{sese_kernel}
    \end{subfigure}
    \caption{Extrapolation of standardized CO$_2$ time series data (yellow dots for training and red dots for test) using GP with three kernels. The dark solid line represents the predictive mean while the shaded area stand for the model confidence. Panel (a) displays the result using single GP with SE kernel. Panel (b) is obtained following the kernel composition in~\cite{rasmussen2006gaussian}. Panel (c) is from using the effective kernel of 2-layer zero-mean DGP with SE used in both layers~\cite{lu2020interpretable}.}\label{fig:kernel_composition}
\end{figure}

Fig.~\ref{fig:kernel_composition}({\bf a}) and ({\bf b}) show fitting and extrapolating the classic carbon dioxide data (yellow marks for training, red for test) with GPs using, respectively, the SE kernel and a mixture of SE, periodic SE and rational quadratic kernels~\cite{rasmussen2006gaussian}, 
\begin{equation*}
	k_{mix}(t,t')=\theta_1^2\exp[-\frac{(t-t')^2}{\theta_2^2}]+\theta_3^2\exp[-\frac{\sin^2(t-t')}{\theta_4^2}]+\theta_5^2[1+\frac{(t-t')^2}{\theta_6^2}]^{-\theta_7}\:.
\end{equation*} All the $\theta$'s are hyperparameters in the mixture kernel.
As a result of the multiple time scales appearing in the data, the vanilla GP fails to capture the short time trend, but, 
the GP with mixture of kernels can still present excellent expressivity and generalization. The log marginal likelihood (logML) is 144 and 459 for the vanilla GP and kernel mixture GP, respectively. 
The two-layer zero-mean DGP with SE kernel in both layers shall perform better than the single-layer counterpart. In Fig.~\ref{fig:kernel_composition}({\bf c}), the GP with the SE[SE] effective kernel 
has excellent fitting with the training data but extrapolates poorly. The good fitting may result from the fact that the SE[SE] kernel does capture the character of multiple length scales in DGP. The logML for the SE[SE] GP is 338.

\begin{figure}[H]
\widefigure
    \centering
    \begin{subfigure}{4.2 cm}
        \includegraphics[width=\textwidth]{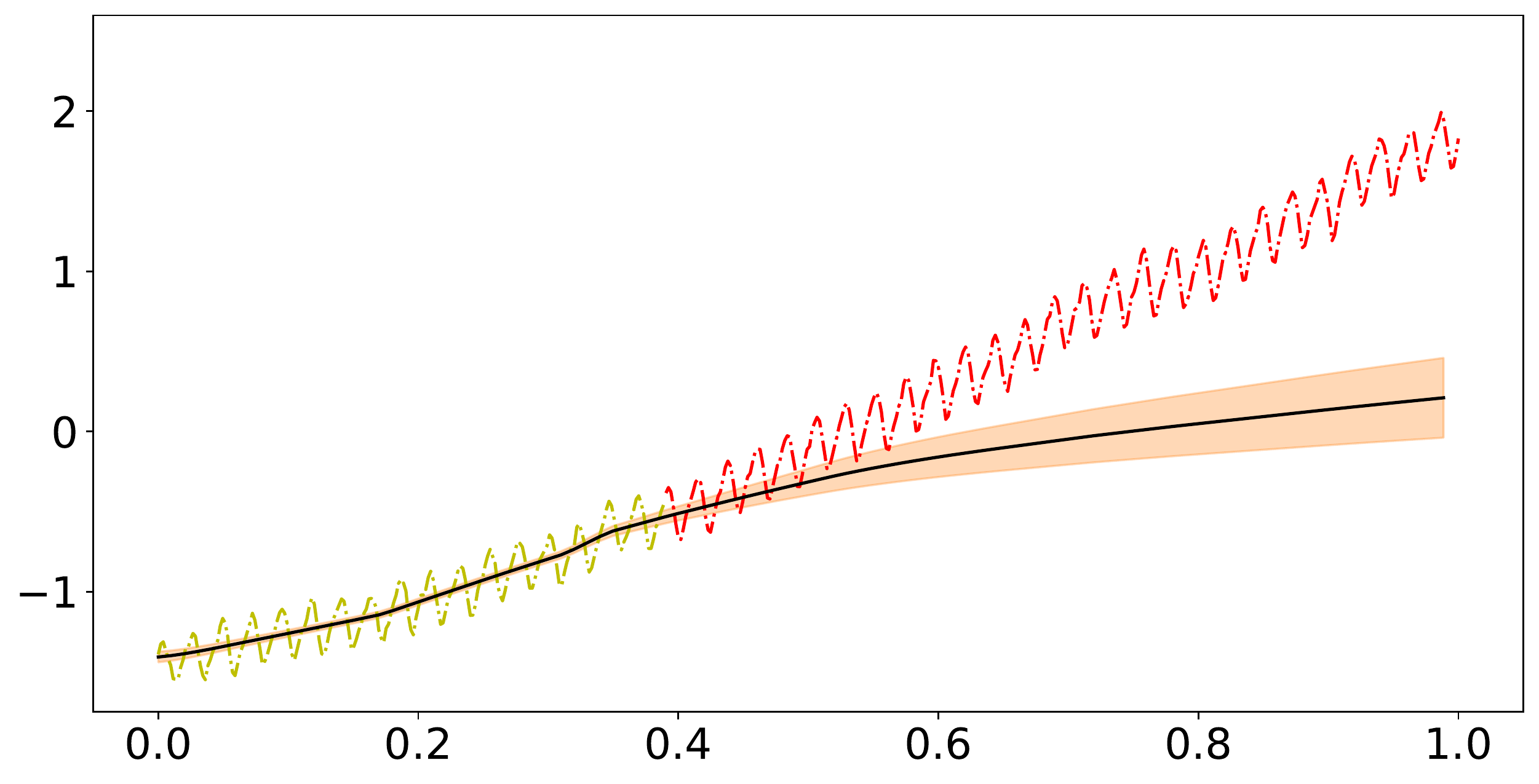}
        \caption{DKL}
        \label{dgp2}
    \end{subfigure}
    ~ 
    \begin{subfigure}{4.2 cm}
        \includegraphics[width=\textwidth]{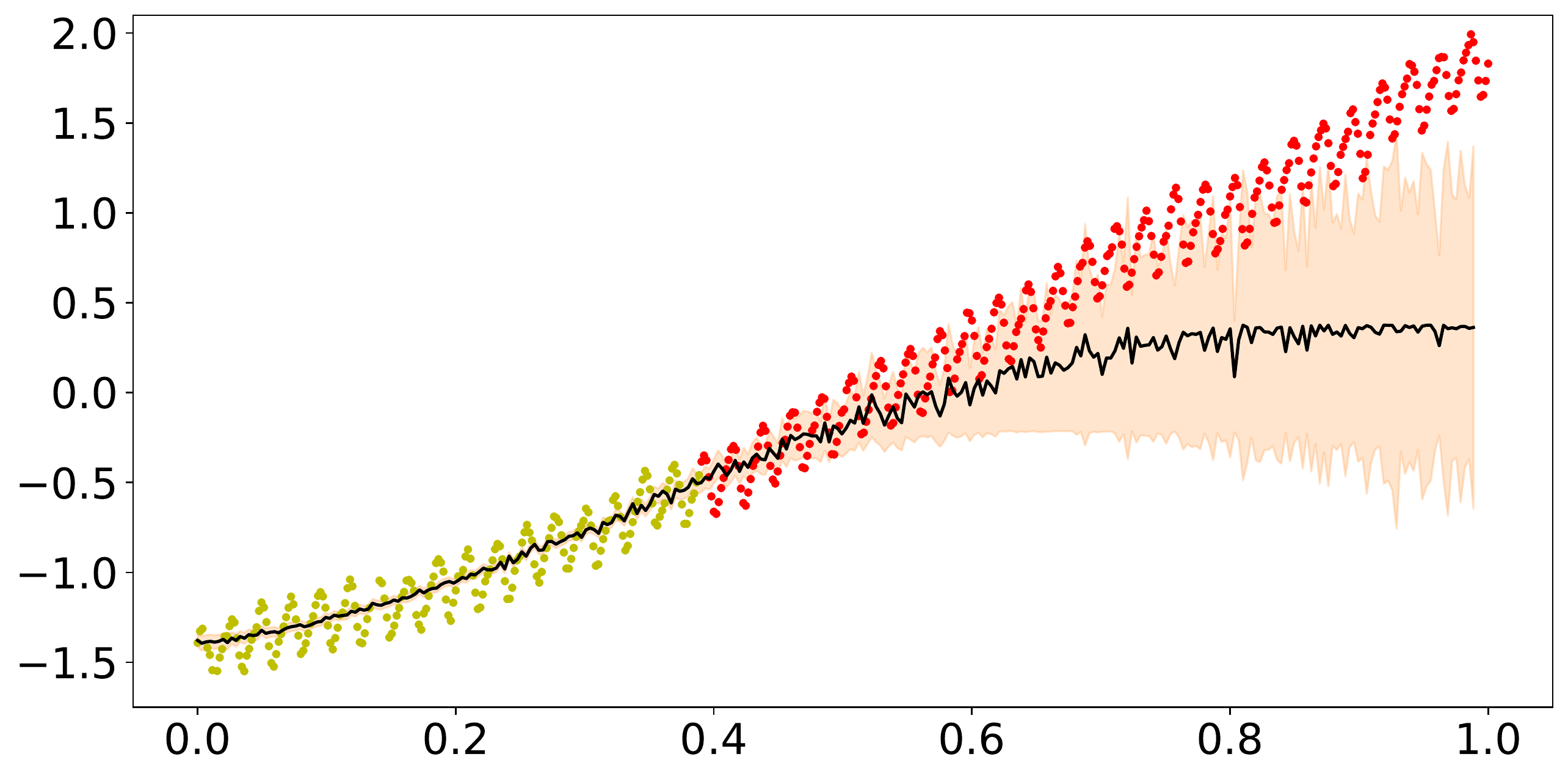}
        \caption{Two-layer DGP}
        \label{dgp3}
    \end{subfigure}
    \begin{subfigure}{4.2 cm}
        \includegraphics[width=\textwidth]{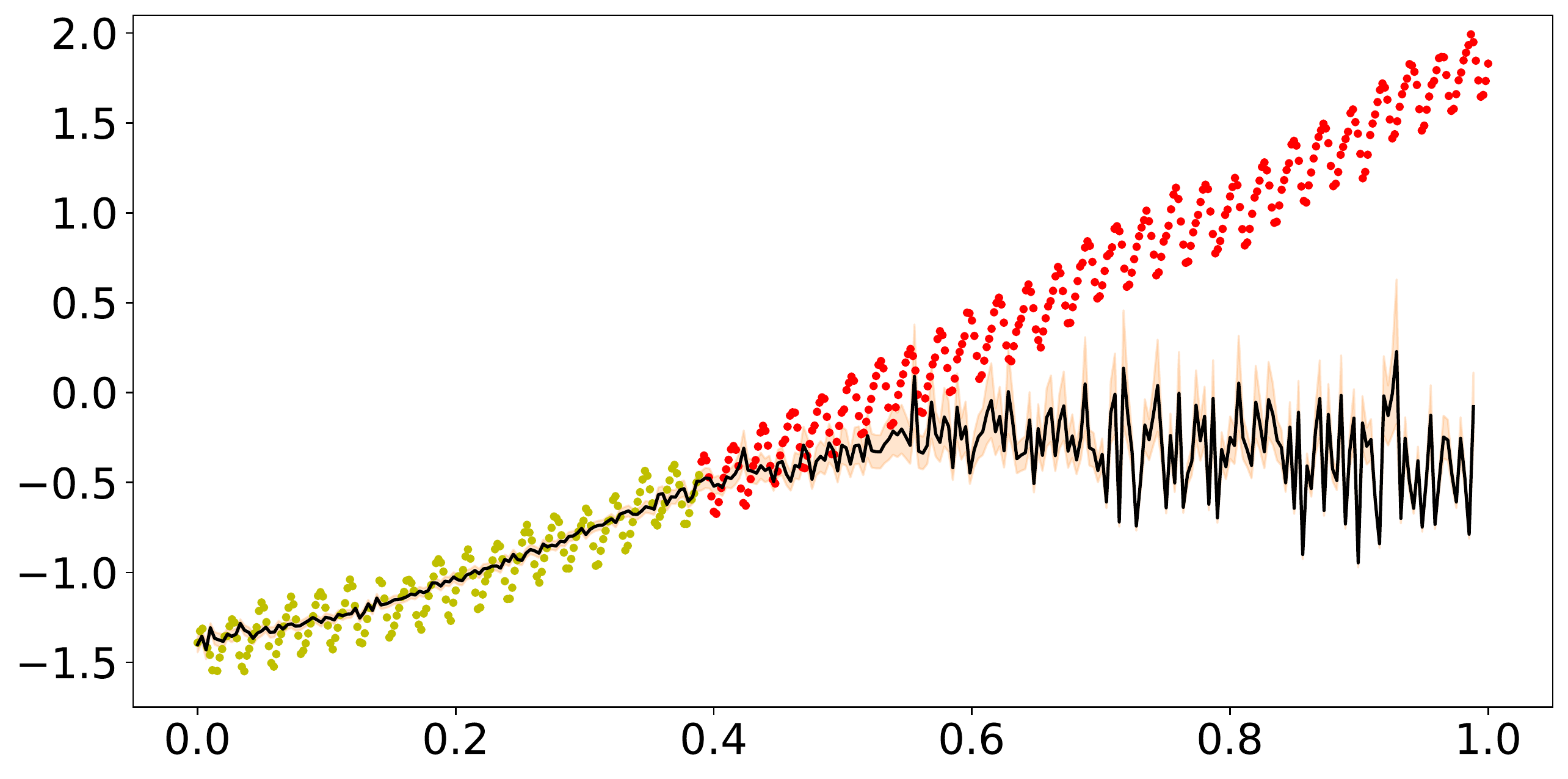}
        \caption{Three-layer DGP}
        \label{dgp3}
    \end{subfigure}
    \caption{Extrapolation of standardized CO$_2$ using DKL and variational inference~\cite{salimbeni2017doubly} for DGP implemented in GPFlux~\cite{dutordoir2021gpflux}. Panel (a) is obtained using the DKL with three-layer RELU network in Fig. Panel (b) is the result from the two-layer zero-mean DGP model, and panel (c) from the three-layer zero-mean DGP.}\label{fig:gpflux}
\end{figure}

Next, we shall see whether an improved extrapolation can arise in other deep models or other inference schemes. In Fig.~\ref{fig:gpflux}, the results from DKL and from DGP using the variational inference are shown. Both are implemented in GPFlux~\cite{dutordoir2021gpflux}. We modified the tutorial code for hybrid GP with three-layer neural network~\hreflink{https://secondmind-labs.github.io/GPflux/notebooks/gpflux_with_keras_layers.html} as the code for DKL. The result in Fig.~\ref{fig:gpflux}({\bf a}) does not show good fitting nor good extrapolation, which is to some extent consistent with the simulation of Bayesian neural network with ReLu activation~\cite{pearce2020expressive}. As for the DGP using variational inference, the deeper models do not show much improvement comparing with the vanilla GP, and the obtained ELBO is 135 for the two-layer DGP [Fig.~\ref{fig:gpflux}({\bf b})] and 127 for three-layer [Fig.~\ref{fig:gpflux}({\bf c})]. 



\begin{figure}[H]
\widefigure
    \centering
    \begin{subfigure}{6.5 cm}
        \includegraphics[width=\textwidth]{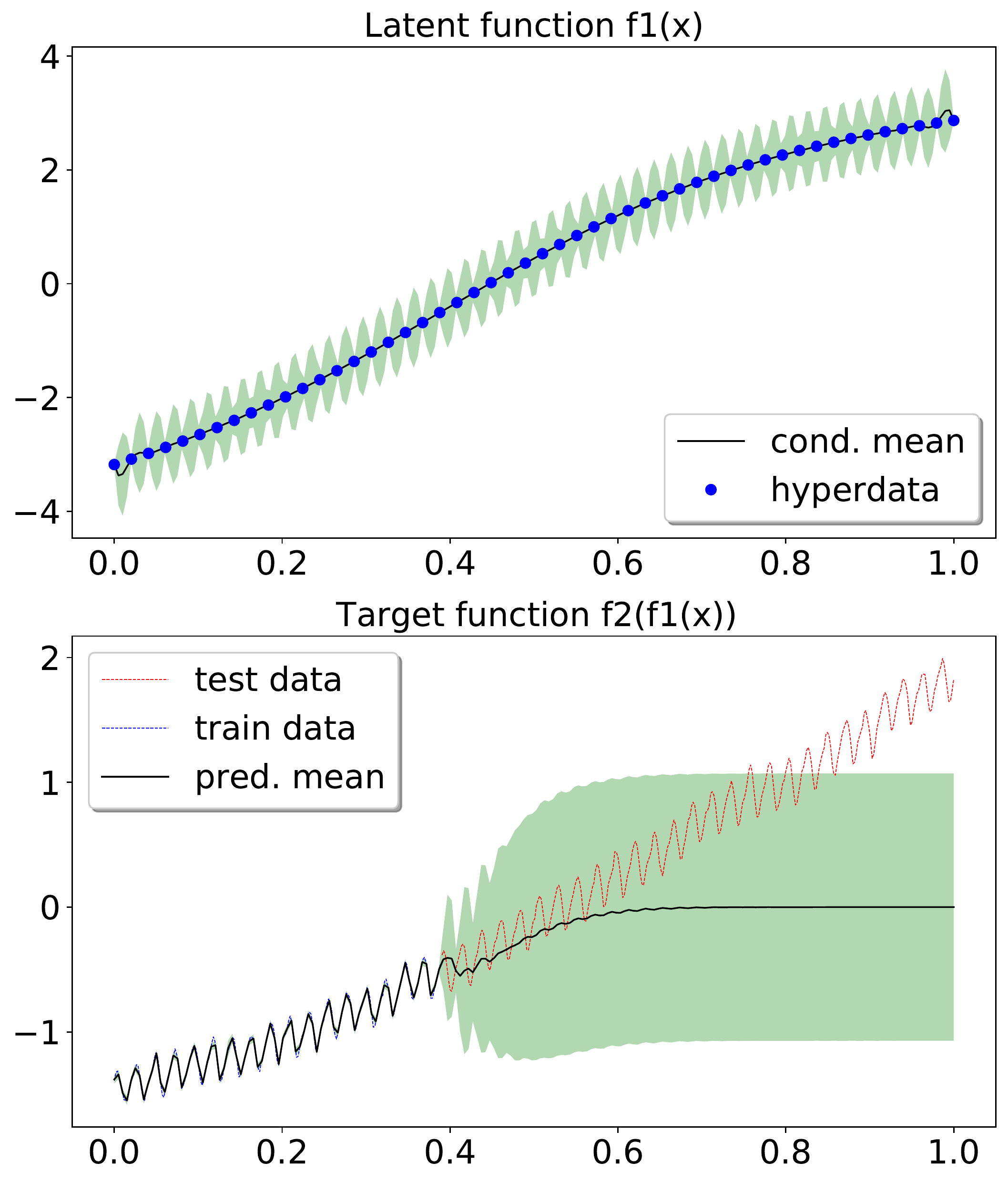}
        \caption{Two-layer conditional DGP}
        \label{dgp1}
    \end{subfigure}
    ~ 
    \begin{subfigure}{6.5 cm}
        \includegraphics[width=\textwidth]{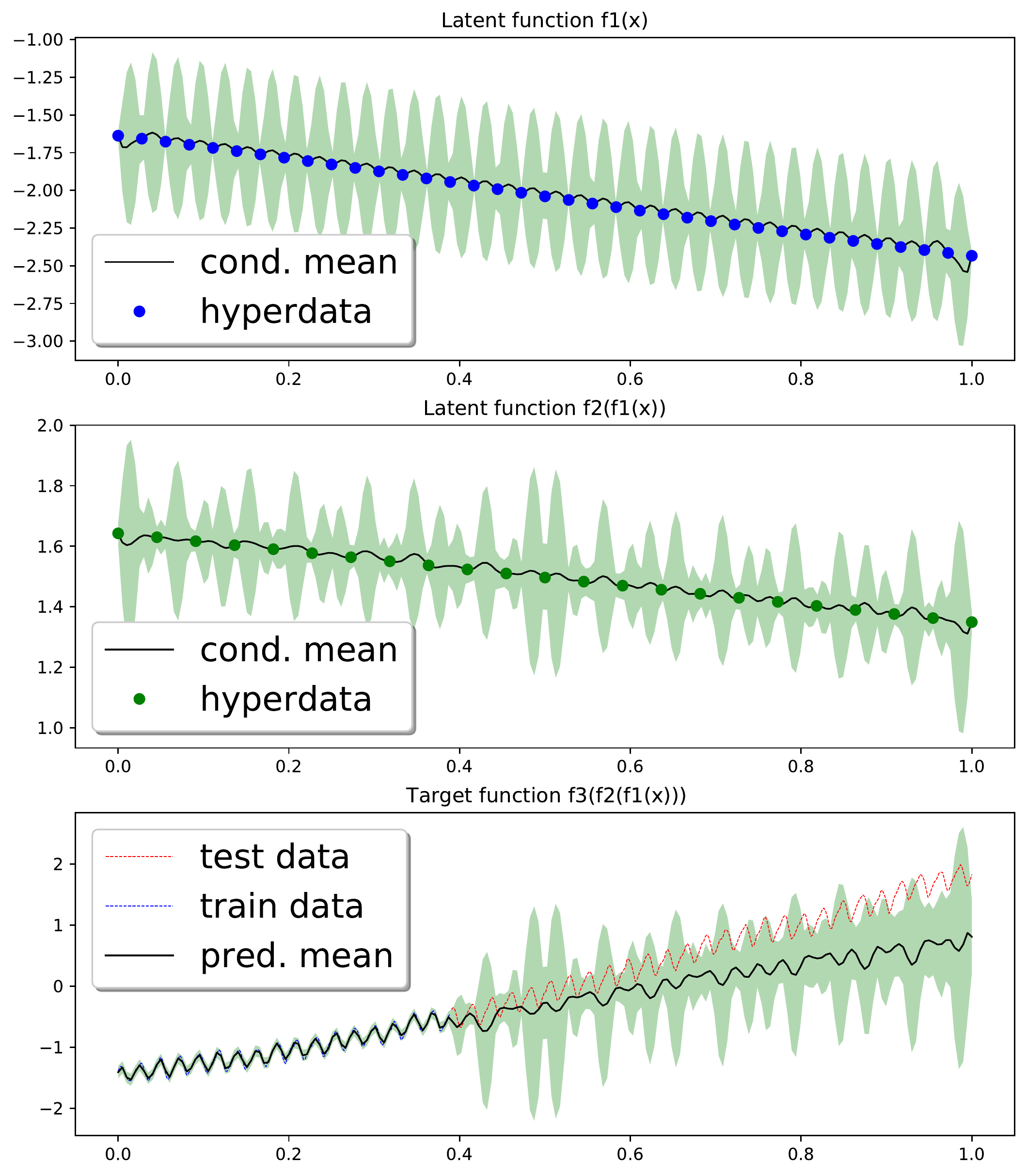}
        \caption{Three-layer conditional DGP}
        \label{dgp2}
    \end{subfigure}
    \caption{Extrapolation of the standardized CO$_2$ using conditional DGP. Panel (a) is for the two-layer model, and (b) for the three-layer model. Top and middle panels shows the mean and confidence in the posterior over the latent functions. See text for details.}\label{fig:co2}
\end{figure}

Now we continue to see the performance of our model. In the two-layer model, we have 50 points in hyperdata supporting the intermediate GP. A width-5 tanh neural network is used to represent the hyperdata, i.e. $u_{1:50}={\rm nn}_{{\bf w}}(z_{1:50})$. Then, the hyperparameters including $\sigma_{1,2}$, $\ell_{1,2}$, and weight parameters ${\bf w}$ are learned from gradient descent upon the approximate marginal likelihood. The top panel in Fig.~\ref{fig:co2}({\bf a}) displays the prediction and confidence from the posterior over $f_1$, obtained from a GP conditioned on the learned hyperparameters and hyperdata. The logML of the two-layer model is also 338, the same as the SE[SE] GP, and in the bottom panel of Fig.~\ref{fig:co2}({\bf a}) one can observe a good fit with the training data. More importantly, the extrapolation shows some high frequency signal in the confidence (shaded region). In comparing with Fig. 3 of~\cite{duvenaud2013structure}, the high-frequency signal only appeared after a periodic kernel is inserted. We attribute the high-frequency signal to the propagation of uncertainty in $f_1$ (top panel) to the exposed layer (see discussion in Sec. 4.2).   

Lastly, the three-layer model using 37 and 23 hyperdata in the $f_1$ and $f_2$ layer, respectively, has the result in Fig~\ref{fig:co2}({\bf b}). Those hyperdata are parameterized by the same neural network used in the two-layer model. The training has a logML of 253, resulting in a good fit with the data. The extrapolation
captures the long term trend in its predictive mean and the test data are mostly covered in the confidence region. In the latent layers, more expressive pattern overlaying the latent mappings seem to emerge due to the uncertainty and the depth of the model. The learned $\sigma_{1,2,3}\approx(0.49,0.86,2.56)$ and $\ell_{1,2,3}\approx(0.014,1.2,0.46)$ shows that different layers manage to learn different resolutions.

\subsection{Airline data}

\begin{figure}[H]
\widefigure
    \centering
    \begin{subfigure}{4.2 cm}
        \includegraphics[width=\textwidth]{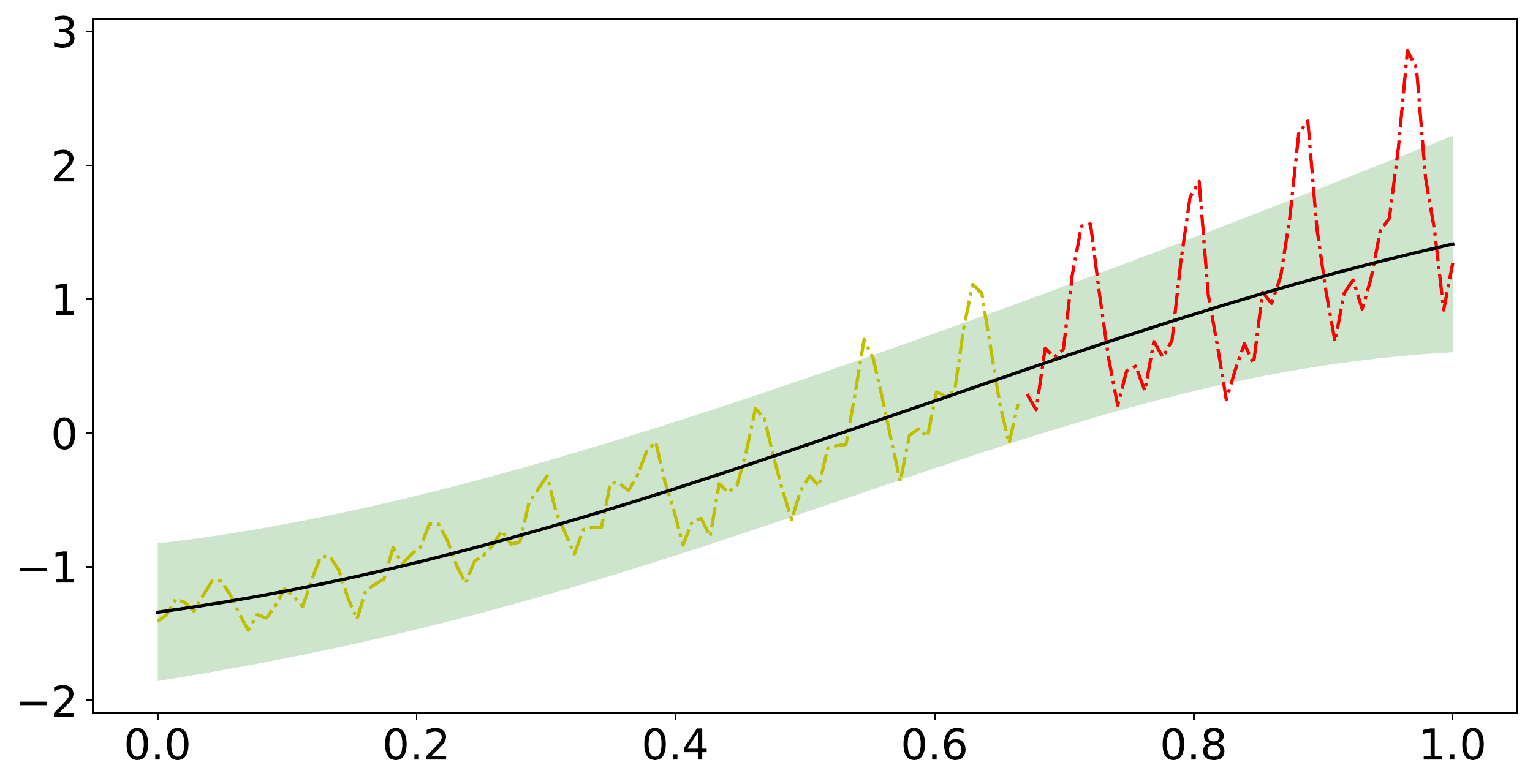}
        \caption{SE kernel}
        \label{se_kernel}
    \end{subfigure}
    ~ 
    \begin{subfigure}{4.2 cm}
        \includegraphics[width=\textwidth]{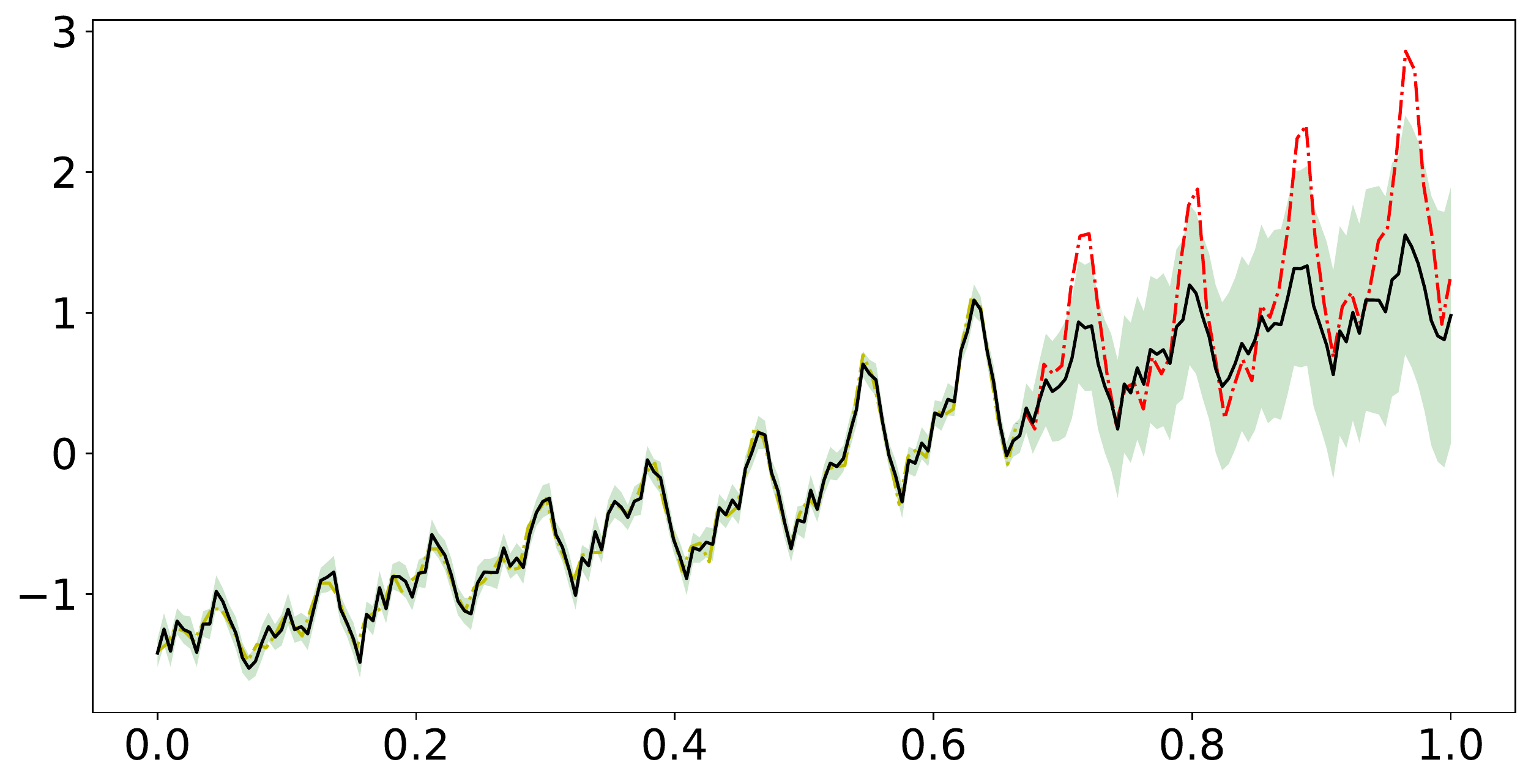}
        \caption{SE+periodic SE+RQ kernel}
        \label{mix_kernel}
    \end{subfigure}
    ~ 
    \begin{subfigure}{4.2 cm}
        \includegraphics[width=\textwidth]{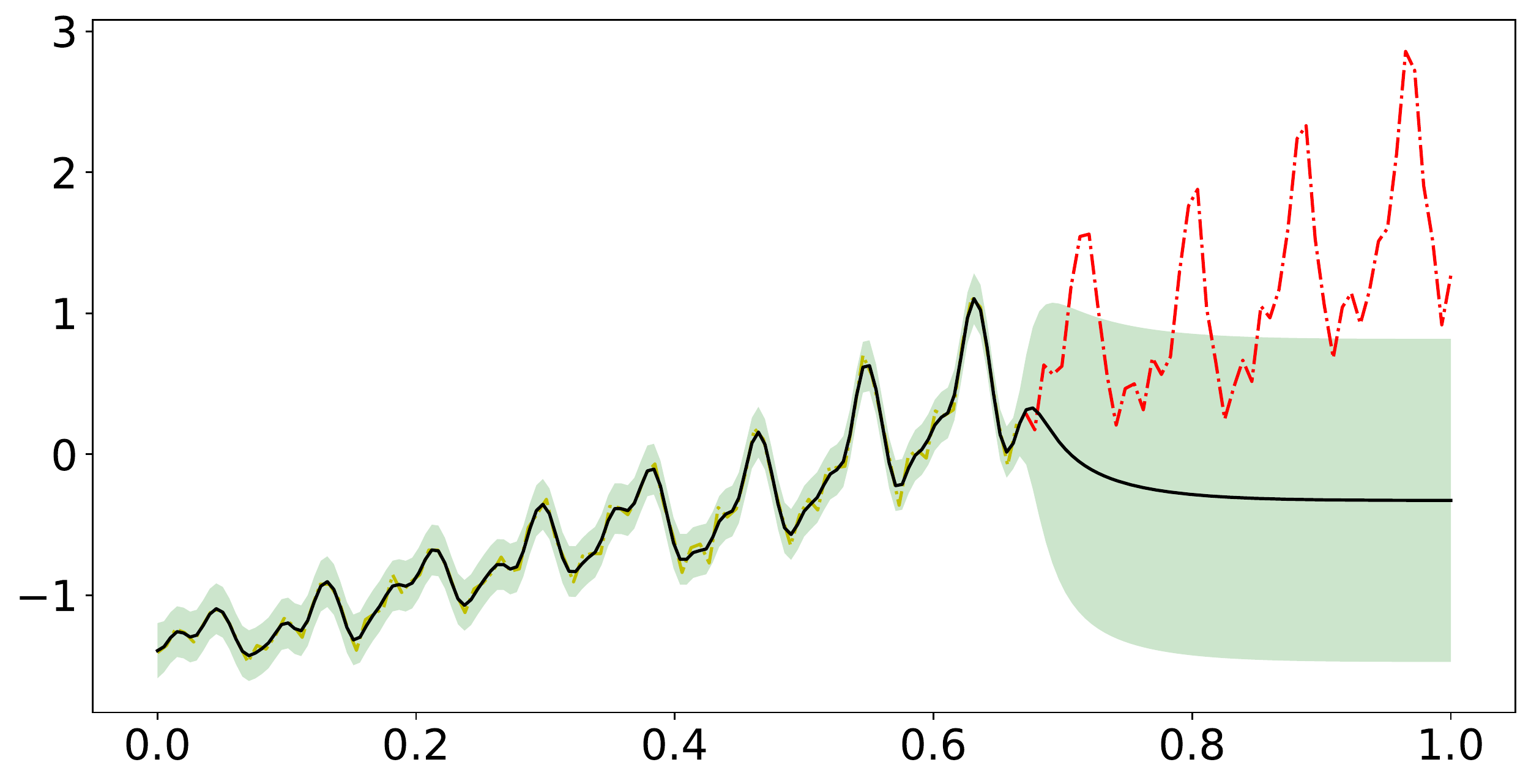}
        \caption{SE[SE] kernel}
        \label{sese_kernel}
    \end{subfigure}
    \caption{Extrapolation of the standardized airline data with three different GPs.}\label{fig:kernel_composition_airline}
\end{figure}

The models under consideration can be applied to the airline data, too. It can be seen in Fig.~\ref{fig:kernel_composition_airline} that the vanilla GP is too simple for the complex time-series data while the GP with the same kernel composition can both fit and extrapolate well. The logML is -11.7 and 81.9 for the vanilla GP and kernel mixture GP, respectively. Similarly, the SE[SE] kernel captures the multiple length scale character in the data, resulting in a good fit with logML 20.9 but a poor extrapolation. 

\begin{figure}[H]
\widefigure
    \centering
    \begin{subfigure}{4.2 cm}
        \includegraphics[width=\textwidth]{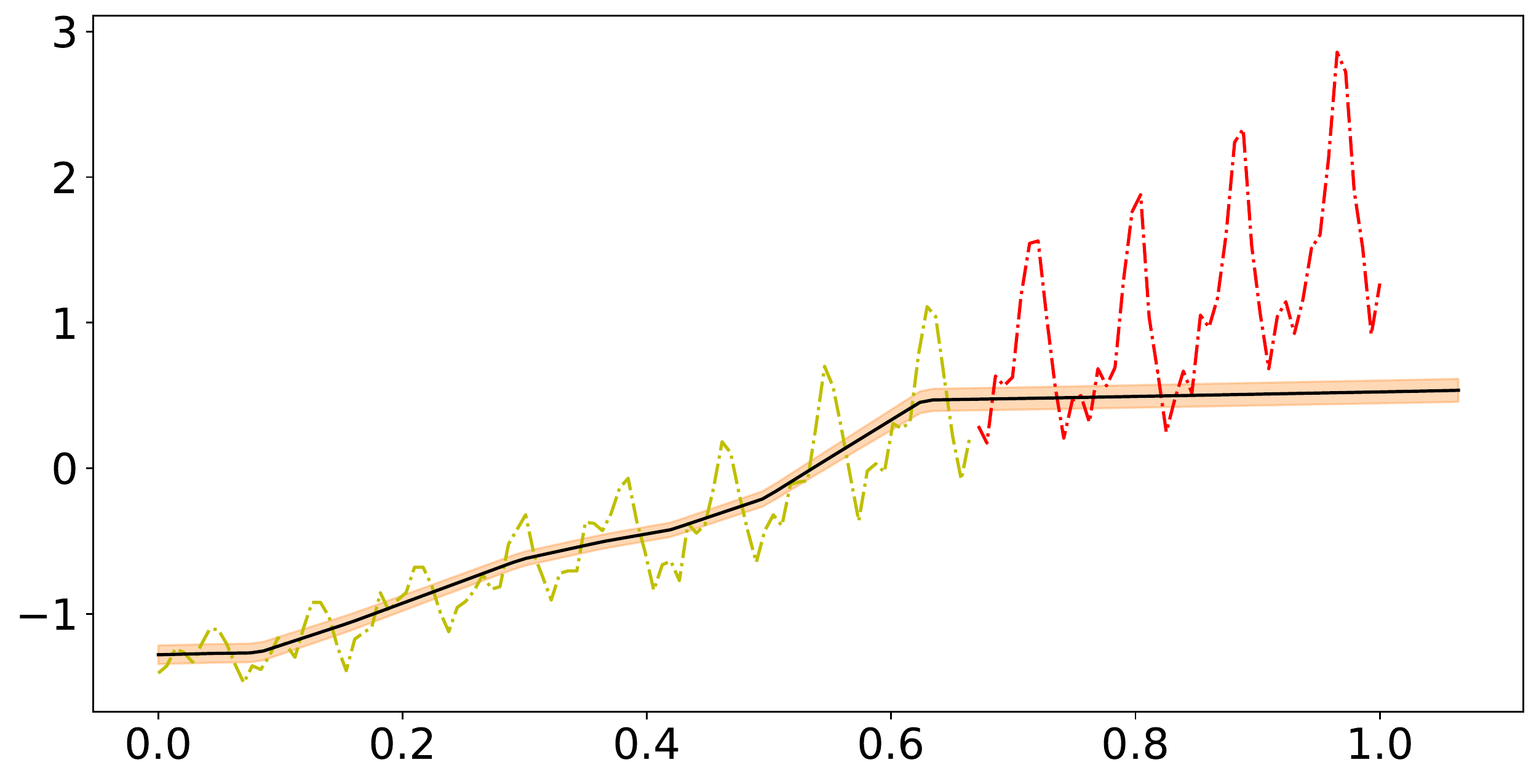}
        \caption{DKL}
        \label{dgp2}
    \end{subfigure}
    \begin{subfigure}{4.2 cm}
        \includegraphics[width=\textwidth]{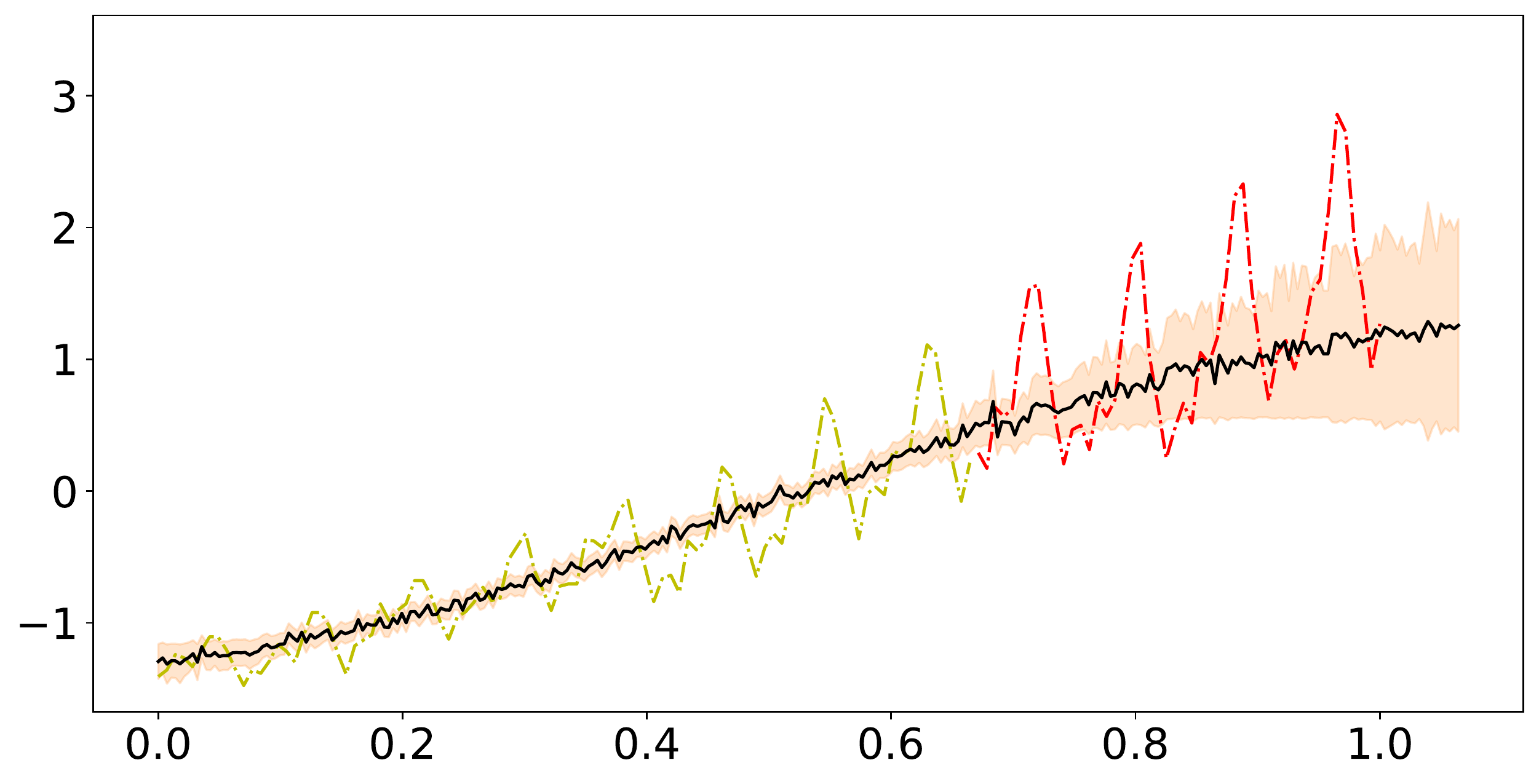}
        \caption{Two-layer DGP}
        \label{dgp3}
    \end{subfigure}
    \begin{subfigure}{4.5 cm}
        \includegraphics[width=\textwidth]{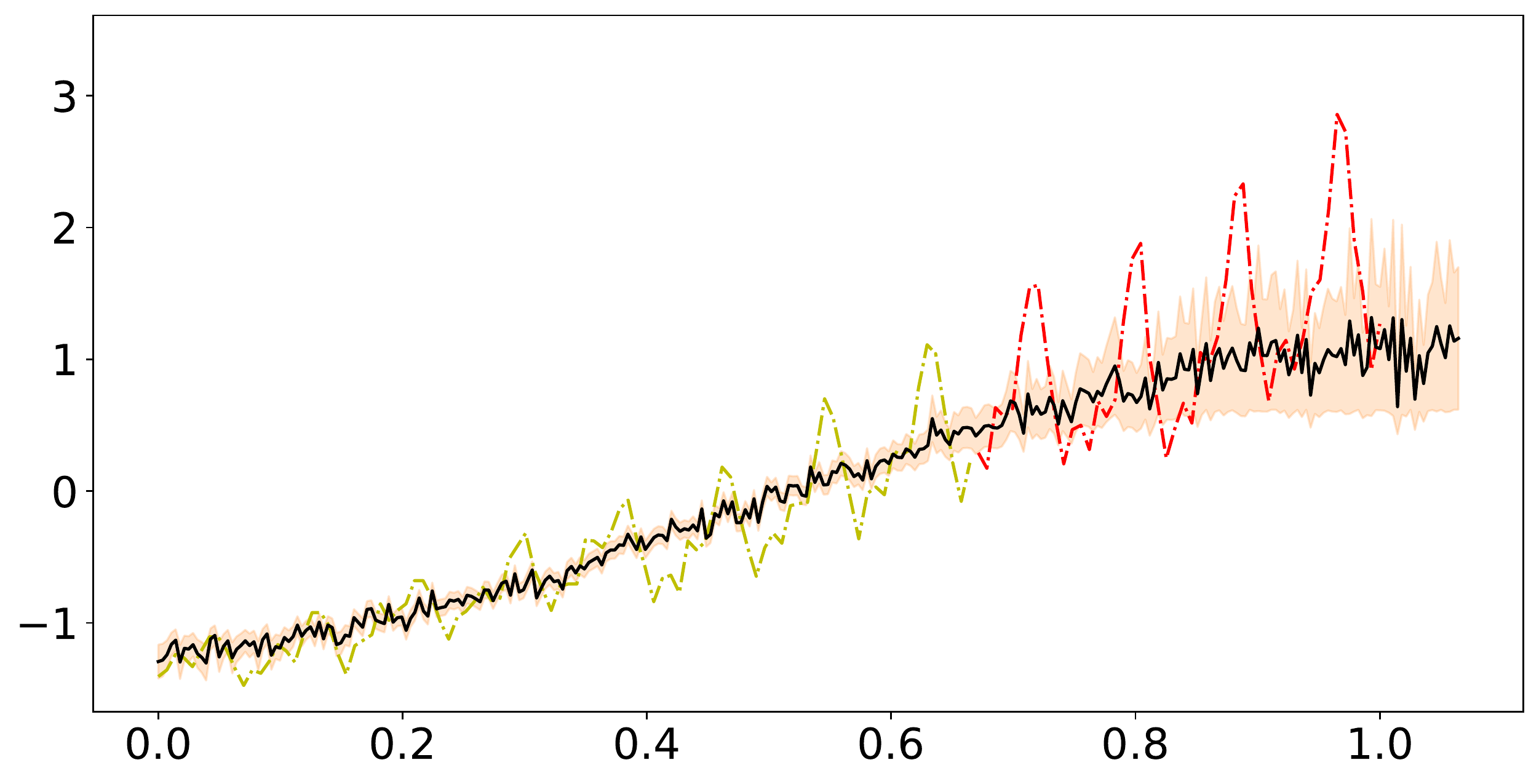}
        \caption{Three-layer DGP}
        \label{dgp3}
    \end{subfigure}
    \caption{Extrapolation of the standardized airline data using DKL (a), 2-layer DGP (b), and 3-layer DGP (c).}\label{fig:gpflux_airline}
\end{figure}

Here, we display the results using the DKL, variational inference DGP, both two-layer and three-layer, in Fig.~\ref{fig:gpflux_airline}. For the airline data, the DKL with ReLu neural network as feature extractor [panel (a)] has similar performance with its counterpart in CO$_2$ data, so does the variational DGP [panels(b,c)]. 

\begin{figure}[H]
\widefigure
    \centering
    \begin{subfigure}{5.9 cm}
        \includegraphics[width=\textwidth]{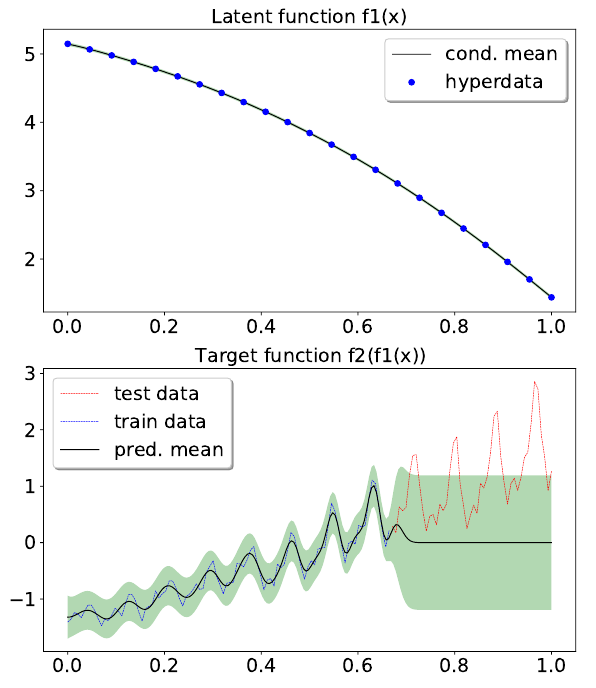}
        \caption{2-layer cDGP with 23 hyperdata}
        \label{cDGP1}
    \end{subfigure}
    \begin{subfigure}{5.9 cm}
        \includegraphics[width=\textwidth]{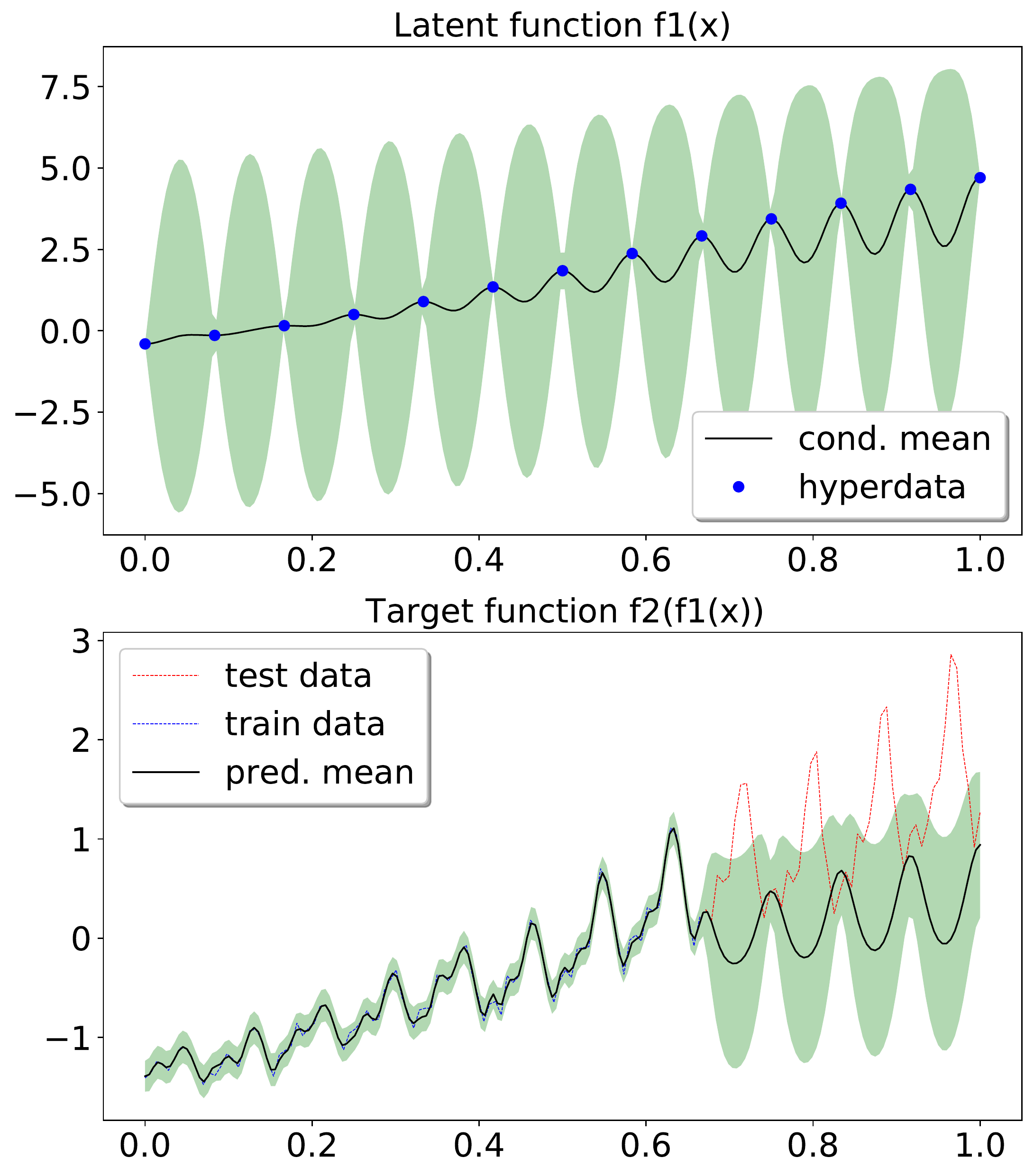}
        \caption{2-layer cDGP with 13 hyperdata}
        \label{cDGP}
    \end{subfigure}
    \caption{Extrapolation of airline data using conditional DGP. The upper panel shows the learned latent function and uncertainty from hyperdata learning, and the bottom panel shows the extrapolation from the past data. Panel (a) has 23 hyperdata supporting the latent GP while panel (b) has 13. Qualitatively, the lack of learned uncertainty in latent function }\label{fig:cdgp_airline}
\end{figure}

Our two-layer model, aided by the probabilistic latent layer supported by 13 hyperdata, shows improved extrapolation along with the high-frequency signal in prediction and confidence. The optimal logML is 28.5, along with the learned $\sigma_{1,2}=(3.03,0.73)$, $\ell_{1,2}=(0.026,2.19)$, and noise level $\sigma_n=0.004$. As shown in Fig.~\ref{fig:cdgp_airline}(b), the latent function supported by learned hyperdata shows the increasing trend on top of an oscillating pattern, which leads to the periodic extrapolation in the predictive distribution albeit only the vanilla kernels are used. It is interesting to compare with Fig.~\ref{fig:cdgp_airline}(a) which have 23 hyperdata supporting the latent function, and the vanishing uncertainty learned in latent function produces the extrapolation collapsed to zero. The logML in panel(a) is 7.25 with learned $\sigma_{1,2}=(2.18,0.59)$, $\ell_{1,2}=(0.3,0.07)$, and noise level $\sigma_n=0.02$.  

\section{Discussion}

What do we gain and lose while modifying the original DGP defined in Eq.~(\ref{true_DGP}) by additionally conditioning the intermediate GPs on the hyperdata? On one hand, when the hyperdata are dense, the conditional DGP is mathematically connected with the deep kernel learning, i.e. a GP with warped input. On the other hand, in the situations when less dense hyperdata present and the latent GPs are representations of random functions passing through the hyperdata, the conditional DGP can be viewed as an ensemble of deep kernels, and the moment matching method allows to express it in a closed form. What do we lose in such approximation? Apparently, the approximate $q$ for the true marginal prior $p$ in Eq.~(\ref{marginal_prior_cond}) can not account for the 
heavy-tailed statistics. 

In the demonstration, the presence of hyperdata constrains the space of the intermediate functions and move the mass of function distribution toward the more probable ones in the process of optimization. Comparing the SE[SE] GP, which represents an approximate version of zero-mean 2-layer DGP, against the conditional DGP model, the constrained space of intermediate functions does not affect the learning significantly while the generalization is improved. Besides, the uncertainty in the latent layers is not collapsed.

One possible criticism of the present model may result from the empirical Bayes learning of the weight parameters. Although the weight parameters are hyperparameter in both our model and in DKL, it is important to distinguish that the weight parameters in our model parameterize $u_{1:M}$, which supports the intermediate GP, representing ensemble of latent functions. In DKL, however, the weight parameters fully determine the one latent function, which might lead to overfitting even though marginal likelihood is used as objective~\cite{ober2021promises}. A possible extension is to consider upgrading the hyperdata to random variables, and the associated mean and variance in $q(u_{1:M})$ can also be modeled as neural network functions of ${\bf Z}$. The moment matching  can then be applied to approximate the marginal prior $\int d{\bf f}_1d{\bf u}p({\bf f}_2|{\bf f}_1)p({\bf f}_1|{\bf X,Z,u})q({\bf u})$.


\section{Conclusions}
Deep Gaussian Processes (DGPs), based on nested composition of Gaussian Processes (GPs), offer the possibility of expressive inference and calibrated uncertainty, but are limited by intractable marginalization. Approximate inference for DGPs via inducing points and variational inference allows scalable inference, but incurs costs in limiting expressiveness and inability to propagate uncertainty. 
We introduce effectively deep kernels with optimizable hyperdata supporting latent GPs via a moment-matching approximation. 
The approach allows joint optimization of hyperdata and GP parameters via maximization of marginal likelihood. 
We show that the approach avoids mode collapse, connects DGPs and deep kernel learning, effectively propagates uncertainty. Future direction on conditional DGP includes consideration of randomness in the hyperdata and the corresponding inference.





\vspace{6pt} 



\authorcontributions{
Conceptualization, C.-K.L. and P.S.; methodology, C.-K.L.; software, C.-K.L.; validation, C.-K.L. and P.S.; formal analysis, C.-K.L.; investigation, C.-K.L.; resources, C.-K.L.; data curation, C.-K.L.; writing---original draft preparation, C.-K.L.; writing---review and editing, C.-K.L. and P.S.; visualization, C.-K.L.; supervision, P.S.; project administration, C.-K.L.; funding acquisition, P.S. All authors have read and agreed to the published version of the manuscript.
}

\funding{
This research was funded by the Air Force Research Laboratory and DARPA under agreement number FA8750-17-2-0146.
}

\conflictsofinterest{The authors declare no conflict of interest. The funders had no role in the design of the study; in the collection, analyses, or interpretation of data; in the writing of the manuscript, or in the decision to publish the~results.} 



\abbreviations{Abbreviations}{
The following abbreviations are used in this manuscript:\\

\noindent 
\begin{tabular}{@{}ll}
GP & Gaussian Process\\
DGP& Deep Gaussian Process\\
DKL& Deep Kernel Learning\\
SE& Squared Exponential\\
\end{tabular}}

\appendixtitles{no} 
\appendixstart
\appendix
\section{}
\subsection{}

\begin{Lemma}
Consider the marginal prior for 2-layer conditional DGP [Eq.~(\ref{marginal_prior_cond})] with $f_2|f_1$ being a GP with SE kernel, and $f_1|{\bf Z,u}$ being another GP with conditional mean $\mu$ and conditional covariance $k$. The general fourth moment is the following sum over distinct doublet decomposition,
\begin{equation*}
     \EX[f({\bf x}_i)f({\bf x}_j)f({\bf x}_m)f({\bf x}_l)]=\sigma_2^4
    \sum\frac{\alpha_{ab,cd}\alpha_{cd,ab}\beta_{ab,cd}}{\sqrt{D_{ab}D_{cd}-V^2_{ab,cd}}}\label{SE_4th}
\end{equation*} with $V_{ab,cd}=(k_{ad}+k_{bc}-k_{ac}-k_{bd})/\ell_2^2$ and $D_{ab}=1+(k_{aa}+k_{bb}-2k_{ab})/\ell_2^2$. Also, the expressions,
\begin{equation*}
    \alpha_{ab,cd}=\exp\left[\frac{-(m_a-m_b)^2}{2\ell_2^2(D_{ab}-V^2_{ab,cd}/D_{cd})}\right]\:,
\end{equation*} and
\begin{equation*}
    \beta_{ab,cd}=\exp\left[-\frac{(m_a-m_b)(m_c-m_d)V_{ab,cd}}{\ell_2^2(D_{ab}D_{cd}-V^2_{ab,cd})}\right]\:.
\end{equation*} 
\end{Lemma}
\begin{proof}
    Denoting the function value $h_a:=f_1({\bf x}_a)$, we can rewrite the product of the covariance function $k_2(h_a,h_b)k_2(h_c,h_d)=\exp(-\frac{[{\bf h}]^T_{ab,cd}\mathbb J_4[{\bf h}]_{ab,cd}}{2})$ where the row vector $[{\bf h}]^T_{ab,cd}=(h_a,h_b,h_c,h_d)$ and the matrix
    \begin{equation*}
        \mathbb J_4 = \left(\begin{matrix}
                         \mathbb J_2 & 0 \\
                        0 & \mathbb J_2 
                        \end{matrix}\right)\:,
    \end{equation*} where $\mathbb J_2$ is the 2-by-2 matrix with ones in the diagonal and minus ones in the off-diagonal. The above zeros stand for 2-by-2 zero matrices in the off-diagonal blocks. The procedure of obtaining expectation value with respect to the 4-variable multivariate Gaussian distribution $\mathcal N([{\bf h}]_{ab,cd}|\mathbb V_4,\mathbb K_4)$ is similar to the previous one in obtaining the second moment. Namely, applying Lemma 2 in~\cite{lu2020interpretable},
    \begin{equation*}
        \EX[k_2(h_a,h_b)k_2(h_c,h_d)]=
        \frac{\exp(-\frac{1}{2}\mathbb V_4^t\mathbb A_4\mathbb V_4)}{\sqrt{I_4+\mathbb K_4\mathbb J_4}}\:,
    \end{equation*} in which the calculation of inverse of 4-by-4 matrix $I_4+\mathbb K_4\mathbb J_4$ and its determinant is quite tedious but tractable. 
\end{proof}

\end{paracol}
\reftitle{References}
\externalbibliography{yes}
\bibliography{reference1.bib}
\end{document}